\documentclass[runningheads,a4]{llncs}
\usepackage{amssymb}
\usepackage{float}
\usepackage{graphicx}
\usepackage{subfigure}
\usepackage{algorithm}
\usepackage{algorithmicx}
\usepackage[noend]{algpseudocode}
\usepackage{amsmath}
\usepackage{CJK}
\usepackage{verbatim}
\usepackage{url}
\begin{document}
\title{Hyper-Parameter Sweep on AlphaZero General}
%
%\titlerunning{Abbreviated paper title}
% If the paper title is too long for the running head, you can set
% an abbreviated paper title here
%
\author{Hui Wang, Michael Emmerich, Mike Preuss, Aske Plaat}
\authorrunning{Hui Wang et al.}
% First names are abbreviated in the running head.
% If there are more than two authors, 'et al.' is used.
%
\institute{Leiden Institute of Advanced Computer Science, Leiden University,\\ Leiden, the Netherlands\\
\email{h.wang.13@liacs.leidenuniv.nl}\\
\url{http://www.cs.leiden.edu}}
\maketitle              % typeset the header of the contribution
\begin{abstract}
Since AlphaGo and AlphaGo Zero have achieved breakground successes in the game of Go, the programs have been  generalized to solve other tasks. Subsequently, AlphaZero was developed to play Go, Chess and Shogi. In the literature, the algorithms are explained  well. However, AlphaZero contains many parameters, and for neither AlphaGo, AlphaGo Zero nor AlphaZero, there is sufficient discussion about how to set parameter values in these algorithms. Therefore, in this paper, we choose 12 parameters in AlphaZero  and evaluate how these parameters contribute to training. We focus on three objectives~(training loss, time cost and playing strength). For each parameter, we train 3 models using 3 different values~(minimum value, default value, maximum value). We use the game of play 6$\times$6 Othello, on the AlphaZeroGeneral open source re-implementation of AlphaZero. Overall, experimental results show that different values can lead to different training results, proving the importance of such a parameter sweep.  We categorize these 12 parameters into time-sensitive parameters and time-friendly parameters. Moreover, through multi-objective analysis, this paper provides an insightful basis for further hyper-parameter optimization.

\end{abstract}

\section{Introduction}
In recent years, many researchers are interested in deep reinforcement learning. The AlphaGo series algorithms~\cite{Silver2016,Silver2017a,Silver2017b} are representative achievements of deep reinforcement learning, which have been greatly promoting the development and application of artificial intelligence technologies. AlphaGo~\cite{Silver2016} achieves the superhuman ability of playing game of Go. AlphaGo applies tree search to evaluate the positions and selects moves from the trained neural networks. However, this neural network is trained from human experts' game playing data. AlphaGo Zero~\cite{Silver2017a} is the update version of AlphaGo and masters game of Go without human knowledge. AlphaGo Zero collects game playing data purely from self-play for training neural network and also evaluates the positions and selects moves from the trained neural network. AlphaZero~\cite{Silver2017b} is a generalized version of AlphaGo Zero and claims a general framework of playing different games~(such as Go, Chess and Shogi) without human knowledge. AlphaZero framework presents the strong adaptability of such a kind of deep reinforcement learning algorithm which combines self-play, neural network and tree search to solve game playing problems. Therefore, inspired by AlphaGo series algorithms, many analysis reviews, applications and optimization methods~\cite{Granter2017,Wang2016,Fu2016} have been proposed and made the deep reinforcement learning be a more and more hot and practical research field.

However, although a lot of research work have been doing based on AlphaGo series approaches and showing the practicalness and adaptiveness in many different application backgrounds~\cite{Tao2016,Zhang2016}. There is insufficient discussion of how to set the parameters in AlphaGo series algorithms. As for an algorithm, apparently, different parameters setting might lead to different results. A proper parameter setting should be found to guarantee the expected capability of the algorithm. In Deep-Mind's work, we can only find some simple and straightforward sentences to give the values of a part of important parameters. Also, few works from others indicate the parameter setting for these algorithms. Therefore, the parameter optimization seems to be necessary. In our work, we choose the most general framework algorithm in aforementioned AlphaGo series algorithms---AlphaZero, to study.

We use a lightweight re-implementation of AlphaZero: AlphaZeroGeneral, \url{https://github.com/suragnair/alpha-zero-general}.

In order to optimize parameters, it is very important to understand the roles~(functions and contributions) of each parameter appeared in the algorithm. Through analyzing the AlphaZeroGeneral framework, in a single iterative loop, it can be divided into three stages: self-play, training neural network and arena comparison. Except for the parameters of designing the structure of neural networks, there are 12 parameters~(see section~\ref{a0gparas}) in AlphaZeroGeneral.

Therefore, in this paper, we intend to sweep 12 parameters by configuring 3 different values~(minimum value, default value and maximum value) to find most promising parameters\footnote{each experiment only observes one parameter, so in a specific experiment, the values of other parameters are set as their own default values.}. In each single run of experiment of training AlphaZeroGeneral to play 6$\times$6 Othello~\cite{Iwata1994}, we change the value of one parameter and keep the values of the rest parameters as default values(see Table~\ref{defaulttab}). We observe 3 objectives~(see section~\ref{a0gtargets}): training loss, elo rating and time cost in each single run. Based on these different objective observations, this paper gives an intuitive view of evaluating contributions of these parameters and provides an insightful basis for further hyper parameter optimization.

%Next, we summarize an optimal parameter setting through trading off targets by minimizing the training loss and time cost and maximizing the elo rating. Then we apply the optimal parameter setting to train the model and compare the targets with the default setting. In addition, in order to present the generalization capability of AlphaZero, we transfer the optimal and default parameter setting to train to play 6$\times$6 ConnectFour and 8$\times$8 Othello.

Our contributions can be summarized as follows:
\begin{enumerate}
\item We give time cost function of the AlphaZeroGeneral algorithm, see Equation \ref{timecostfunction}.

\item We sweep 12 parameters in AlphaZeroGeneral and provide detailed training results and analysis about loss, elo rating and time cost, respectively for every parameters, see Fig.~\ref{fig:figalldefault}--\ref{fig:subfigupdateThreshold} and Table.~\ref{timecosttab}.

\item We summarize most promising parameters based on different objective for further optimization, see Table.~\ref{losstimetab}.
\end{enumerate}

\section{Related work}

Parameter tuning by optimization is very important for many practical algorithms. In reinforcement learning, for instance, the $\epsilon$-greedy strategy of classical Q-learning is used to balance the exploration and exploitation. Different $\epsilon$ values lead to different learning performance~\cite{Wang2018}. Another well known example of parameter tuning in reinforcement learning is the parameter $C_p$ in UCT~(Upper Confidence Bound Apply to Tree) formula~\cite{Browne2012}. There are a lot of work to tune $C_p$ for different kinds of tasks and provide a lot of insight of setting its value for Monte Carlo Tree Search~(MCTS) to balance the exploration and exploitation~\cite{Ruijl2014}. In deep reinforcement learning, neural network shows black-box but strong capability~\cite{Schmidhuber2015} of training, such as training convolutional neural network to play Go~\cite{Clark2015}. Since Mnih et al. reported their work on human-level control through deep reinforcement learning~\cite{Mnih2015} in 2015, the performance of deep Q-network~(DQN) shows us an amazing impression on playing Atari 2600 Games. Thereafter, the applications based on DQN have shown surprising ability of learning to solve game playing problems. For example, Silver et al.~\cite{Silver2016} applied tree search and neural network to train and enhance training model to play game of Go based on human experts' playing data. Then, based on the research of self-play in reinforcement learning~\cite{Heinz2000,Wiering2010,Van2013}, Silver et al. continued to use self-play to generate training data instead of training from human data, which saves a lot of work of collecting and labeling data from human experts~\cite{Silver2017a}. Soon after, Silver et al.~\cite{Silver2017b} generalized their approach as a framework for dealing with more general tasks~(such as Go, Shogi and Chess in their work). Although there are so many impressive achievements of AlphaGo series algorithms, few works discuss on parameter tuning of these algorithms. Therefore, it is necessary to optimize parameters for these algorithms. Instead of optimizing directly, since there are too many parameters in these algorithms, and we do not know if there are any correlations among these parameters, in our paper, as the first step, filtering for the most promising parameters in order to further optimize these is pursued as an alternative.

\section{AlphaZero}
\subsection{AlphaZero Algorithm}\label{a0gintroduction}
According to~\cite{Silver2017b}, the structure of AlphaZero algorithm is an iteratively loop as a whole, which can be divided into three stages within the loop. During a pure \emph{iteration}, the first stage is \textbf{self-play}. The player plays several games against to itself to generate games data for further training. In each step of a game~(\emph{episode}), the player runs Monte Carlo Tree Search~(MCTS) to obtain the playing policy~($\overrightarrow{pi}$). In MCTS, \emph{Cpuct} is used to balance the exploration and exploitation of game tree search. \emph{mctssimulation} is the number of searching times from the root node to build the game tree, where the current best neural network model~(\emph{nnm}) provides the value of the states for MCTS. Once the player get the $\overrightarrow{pi}$ based on MCTS, in the game, the player always chooses the best move according to $\overrightarrow{pi}$ after \emph{tempThreshold} steps, before \emph{tempThreshold} steps, the player always chooses a random move based on the probability distribution from $\overrightarrow{pi}$. These games examples~(normalized as a form of ($s_t, \overrightarrow{\pi}_t, z_t$)) are appended to the \emph{trainingExamplesList}. If the iteration count surpasses the \emph{retrainlength}, the first iterations examples in the queue will be popped out. The second stage is \textbf{training neural network} using games data from self-play. While training, it is not enough to pull through a training examples set into neural network only once, and it is not possible to pull through the whole training examples set at once time. Therefore, several \emph{epochs} are needed. In each \emph{epoch}, training examples are divided into several small batches according to the specific \emph{batchsize}. Neural network is trained to optimize~(minimize)~\cite{Kingma2014} the value of \emph{loss function}. The last stage is comparing the newly trained neural network model with the previous neural network model~(\textbf{arena comparison}), the player will adopt the better model for the next iteration. In order to achieve this, newly trained neural network model~(\emph{nnnw}) and previous best neural network model~(\emph{pnnw}) are compared in an arena. In the arena, players should pit against to the opponent for \emph{arenacompare} games. If the \emph{nnnw} wins more than a proportion of \emph{updateThreshold}, the \emph{nnnw} will be accepted to replace the \emph{pnnw}. Otherwise, the \emph{nnnw} should be rejected and the \emph{pnnw} will still be used as current best neural network model. In order to present this process intuitively, we introduce the pseudo code as Algorithm~\ref{alg:a0g}, where all 12 parameters are presented in their corresponding positions.
\allowdisplaybreaks
\begin{algorithm}[tbh]
%\footnotesize
\caption{AlphaZero Algorithm}
\label{alg:a0g}
\begin{algorithmic}[1]
\Function{AlphaZero}{initial neural network model: nnm, parameter setting: ps}
\While{current iteration$<$ps.iteration}\label{selfplaystarts} \Comment{stage 1}
\While{current episode$<$ps.episode}
\While{!game terminates}
\State $\overrightarrow{\pi}\leftarrow$ MCTS(nnm, ps.Cpuct, ps.mctssimulation, $s$);
\If{current game step$<$ps.tempThreshold}
\State action$\sim \overrightarrow{\pi}$;
\Else
\State action$\leftarrow \arg\max_a{\overrightarrow{\pi}}$;
\EndIf
\EndWhile
\EndWhile
\State trainingExamplesList$\leftarrow$ append examples;
\If{iteration of trainingExamplesList$>$ps.retrainlength}
\State trainingExamplesList.pop(0);\label{selfplayends}
\EndIf
\While{current epoch$<$ps.epoch}\Comment{stage 2}
\State batch$\leftarrow$ trainingExamples.size/ps.batchsize;
\While{current batch$<$batch}
\State loss\_pi, loss\_v, nnnw$\leftarrow$ trainNNT( ps.learningrate, ps.dropout, training examples of current batch) by optimizing the loss function;
%\State update loss\_pi, loss\_v;
\EndWhile
\EndWhile
\While{current arenacompare$<$ps.arenacompare}\Comment{stage 3}
\While{!game terminates}
\If{player==player1}
\State $\overrightarrow{\pi_1}\leftarrow$ MCTS(nnnm, ps.Cpuct, ps.mctssimulation, $s$);
\State action$\leftarrow \arg\max_a{\overrightarrow{\pi_1}}$;
\Else
\State $\overrightarrow{\pi_2}\leftarrow$ MCTS(pnnm, ps.Cpuct, ps.mctssimulation, $s$);
\State action$\leftarrow \arg\max_a{\overrightarrow{\pi_2}}$;
\EndIf
\EndWhile
\EndWhile
\If{player1.win/ps.arenacompare$\geq$ps.updateThreshold}
\State nnm$\leftarrow$nnnm;
\EndIf
\EndWhile
\State \Return nnm;
\EndFunction
\end{algorithmic}
\end{algorithm}
\subsection{Parameters in AlphaZero Algorithm}\label{a0gparas}

In order to make it clear, in this paper, parameters are presented as the order of their first showing up order in the Algorithm~\ref{alg:a0g}.

\noindent{\textbf{\emph{iteration}} is used to set the number of iteration for the whole training process.}

\noindent{\textbf{\emph{episode}} is used to set the number of episode for self-play. One episode means a whole circle from game start to the end.}

\noindent{\textbf{\emph{tempThreshold}} is used to judge that the player should take the best action or not. In each game~(episode), if the game step is smaller than \emph{tempThreshold}, the player will choose an action randomly according to the probability distribution of all legal actions, otherwise, the player will always choose the best action.}

\noindent{\textbf{\emph{mctssimu}} is used to set the number of MCTS Simulation. This number controls the building of search tree according to the neural network.}

\noindent{\textbf{\emph{Cpuct}} is used in UCT formula, which is very useful to balance the exploration and exploitation in the MCTS.}

\noindent{\textbf{\emph{retrainlength}} is used to set the number of last iterations. The neural network will also be trained using training examples from these iterations again and again. Except from the training examples of the current iteration, other training examples are trained one or more times, so this is called retrain. Retrain is used to avoid over fitting while training the neural network.}

\noindent{\textbf{\emph{epoch}} is used to set the number of epoch. It is called one epoch while all training examples pass through the neural network. Usually, one epoch is not enough this is because too few epoch may lead under fitting. However, too many epochs, in the contrast, will lead over fitting.}

\noindent{\textbf{\emph{batchsize}} is used to set the number of size of the training example batch. It is always impossible to make all training examples pass through the Neural Network at one time, so dividing the training examples into smaller batch is an applicable way~\cite{Ioffe2015}.}

\noindent{\textbf{\emph{learningrate}} is used to control the learning rate of the neural network model.}

\noindent{\textbf{\emph{dropout}} is used to set the probability. The neural network will ignore some nodes in the hidden layer randomly based on this probability. Such mechanism is also used to reduce over fitting~\cite{Srivastava2014}.}

\noindent{\textbf{\emph{arenacompare}} is used to set the number of games in arena to compare the new neural network model with the old one.}

\noindent{\textbf{\emph{updateThreshold}} is used to determine that the training process should accept the new neural network model or not. In the arena, if the win rate of the new neural network is higher than \emph{updateThreshold}, the training process will accept the new neural network model as the temporary best model.}

%\noindent{\textbf{\emph{channel}} is used to set the number of stencil channel of Convolutional Neural Network}

\subsection{Objectives}\label{a0gtargets}
In our design, we observe training loss, elo rating and time cost for the learning player based on Algorithm~\ref{alg:a0g}.

\noindent{\textbf{\emph{training loss function}} consists of policy loss~(\emph{loss\_pi}) and value loss~(\emph{loss\_v}). The neural network $f_{\theta}$ is parameterized by $\theta$. $f_{\theta}$ takes the game board state $s$ as input, and the value $v_{\theta}\in [-1,1]$ of $s$ and a policy probability distribution vector $\overrightarrow{pi}$ over all legal actions as outputs. $\overrightarrow{pi}_{\theta}$ is the policy provided by $f_{\theta}$ to guide MCTS for playing games. After performing MCTS, there is an improvement estimate $\overrightarrow{\pi}$, one of training targets is to make $\overrightarrow{\pi}$ be more and more similar as $\overrightarrow{pi}$. This can be achieved by minimizing the cross entropy of two distributions. Therefore, the \emph{loss\_pi} can be defined as $-\overrightarrow{\pi}\log(\overrightarrow{pi}_{\theta}(s_t))$. The other target of training neural network is to minimizing the difference between the output value~($v_{\theta}(s_t)$) of the $s$ according to $f_{\theta}$ and the real outcome~($z_t\in\{-1,1\}$) of the game. This can be achieved by minimizing the variances of two output values. Therefore, \emph{loss\_v} can be defined as $(v_{\theta}(s_t)-z_t)^2$. Overall, the total loss function can be defined as equation~\ref{totalloss}
\begin{equation}\label{totalloss}
loss\_pi + loss\_v=-\overrightarrow{\pi}\log(\overrightarrow{pi}_{\theta}(s_t))+(v_{\theta}(s_t)-z_t)^2
\end{equation}

\noindent{\textbf{\emph{elo rating function}} is developed as a method for calculating the relative skill levels of players in games~\cite{Albers2001}. Following~\cite{Silver2017b}, in our design, we also adopt bayesian elo system~\cite{Coulom2008} to compute the elo increasing curve of the learning player during the whole training iterations. In zero-sum games, there are two players, player A and B. If player A has an elo rating of $R_A$ and B has an elo rating of $R_B$, then the expectation of that player A wins the next game can be calculated by $E_A=\frac{1}{1+10^{(R_B-R_A)/400}}$. If the real outcome of the next game is $S_A$, then the updated elo rating of player A can be calculated by $R_A=R_A+K (S_A-E_A)$, where K is the factor of the maximum possible adjustment per game. In practice, K should be set as a bigger value for weaker players but a smaller value for stronger players.

\noindent{\textbf{\emph{time cost function}} is a prediction function we summarized from Algorithm~\ref{alg:a0g}. According to Algorithm~\ref{alg:a0g}, the whole training process consists of several iterations. In every single iteration, there are all three steps as we introduced in section~\ref{a0gintroduction}. For self-play, there are three loops. The outer one is for playing episodes, the middle one is the game steps, the inner one is for MCTS simulations in each step. For neural network training, there are also two loops. The outer one is for epochs, the inner one is for batches.  For arena comparison, it is similar as self-play, but outer loop invariance is called arenacompare. Overall, in $i$th iteration~($1\leq i\leq iteration$), if we assume that in $j$th episode~($1\leq j\leq episode$), for $k$th game step~(the size of $k$ mainly depends on the game complexity), the time cost of $l$th MCTS~($1\leq l\leq mctssimulation$) simulation is $t_{jkl}^{(i)}$, and assume that for $u$th epoch~($1\leq u \leq epoch$), the time cost of pulling $v$th batch~($1\leq v \leq trainingExampleList.size/batchsize$\footnote{the size of \emph{trainingExampleList} is also relative to the game complexity}) through the neural network is $t_{uv}^{(i)}$, and assume that in $w$th arena comparison~($1\leq w \leq arenacompare$), for $x$th game step, the time cost of $y$th  MCTS simulation~($1\leq y \leq mctssimulation$) is $t_{xyw}^{(i)}$, the time cost of whole training process can be summarized equation~\ref{timecostfunction}
\begin{equation}\label{timecostfunction}
%t=\sum_{iter=1}^{iter=iteration}(\sum_{epi=1}^{episode}\sum_{simu=1}^{simu=mctssimulation}episimutime_{jk}+\sum_{epo=1}^{epo=epoch}\sum_{bat=1}^{bat=batchsize}battime_{mn}+\sum_{com=1}^{com=arenacompare}\sum_{simu=1}^{simu=mctssimulation}comsimutime_{xy})
t=\sum_{i}(
\stackrel{self-play}{\overbrace{\sum_{j}\sum_{k}\sum_{l}t_{jkl}^{(i)}}}\ +\stackrel{train\ neural\ network}{\overbrace{\sum_{u}\sum_{v}t_{uv}^{(i)}}}+\ \stackrel{arena\ comparison}{\overbrace{\sum_{x}\sum_{y}\sum_{w}t_{xyw}^{(i)}}})
\end{equation}

From Algorithm~\ref{alg:a0g} and equation~\ref{timecostfunction}, we can easily know that the parameters, such as \emph{iteration}, \emph{episode}, \emph{mctssimulation}, \emph{epoch}, \emph{batchsize},  \emph{retrainlength}, \emph{arenacompare} etc., will obviously influence the time cost of training. Besides, $t_{jkl}^{(i)}$ and $t_{xyw}^{(i)}$ are both the simulation time cost, they rely on the hardware capacity. $t_{uv}^{(i)}$ also relies on the structure of neural network. In our experiments, all neural network models share the same structure, which consists of 4 convolutional neural network and 2 fully connected layers.

\section{Set Up}
Our experiments are run on one of our college server, which has 128G RAM, 3TB local storage, 20 Intel Xeon E5-2650v3 CPUs~(2.30GHz, 40 threads), 2 NVIDIA Titanium GPUs~(each with 12GB memory) and 6 NVIDIA GTX 980 Ti GPUs~(each with 6GB memory). In order to keep using the same GPUs, we deploy each run of experiments on the NVIDIA GTX 980 Ti GPU.

Othello~(also called Reversi) is a classic board game and usually played on the 8$\times$8 board. In our approach, first, in order to save time, we sweep 12 parameters introduced in section~\ref{a0gparas} by training to play 6$\times$6 Othello. In each run of our experiments, we only observe one of these parameters. Since each run of experiments is time-consuming, we only set three different values~(called minimum value, default value and maximum value, respectively) for each parameter. As a baseline, we set all parameters as their own default values to run experiments. For example, if we observe \emph{Cpuct}, for instance, we set \emph{Cpuct} as minimum value for the first run~(then maximum value for the second run) and keep all other parameters values as their default value to run experiments. All necessary data generated during experiments, such as loss value, arenacompare results and time cost, should be recorded properly.

In order to train a player to play Othello based on Algorithm~\ref{alg:a0g}, we set values for each parameters in Table~\ref{defaulttab}.
\begin{table}[H]
\centering\hspace*{-2.3em}
%\linebreak
\caption{Default Parameter Setting}\label{defaulttab}
\begin{tabular}{|l|l|l|l|}
\hline
Parameter	& Minimum Value	& Default Value& Maximum Value\\
\hline
iteration	&50	&100	&150\\
\hline
episode	&10	&50&100\\
\hline
tempThreshold	&10		&15&20\\
\hline
mctssimu	&25	&100&	200\\
\hline
Cpuct	&0.5	&1.0&	2.0\\
\hline
retrainlength	&1	&20&40\\
\hline
epoch	&5	&10	&15\\
\hline
batchsize	&32	&64&96\\
\hline
learningrate	&0.001&	0.005&0.01\\
\hline
dropout&	0.2&0.3&0.4\\
%\hline
%channel&	32&	256&512\\
\hline
arenacompare	&20	&40	&100\\
\hline
updateThreshold	&0.5	&0.6 &	0.7\\
\hline
\end{tabular}
\end{table}

\section{Experiment Results}
In order to show the training process clearly, first, we present the results of default setting as Fig.~\ref{fig:figalldefault}.

\begin{figure}[H]
\centering
\hspace*{-2.3em}
\includegraphics[width=0.8\textwidth]{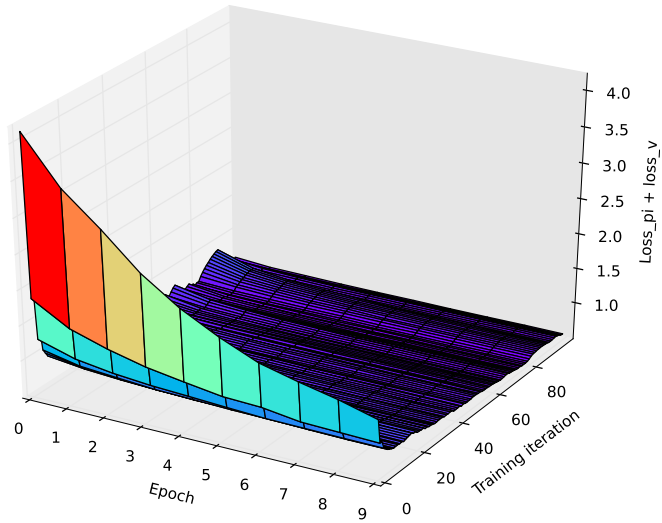}
\caption{Single Run 3D Training Loss}
\label{fig:figalldefault} %% label for entire figure
\end{figure}

In Fig.~\ref{fig:figalldefault}, we plot the training loss of each epoch in every iteration. The results show that, with reasonable fluctuation, (1) in each iteration, the training loss decreases along with the increasing of epoch, and that (2) the training loss also decreases with the increasing of iteration until to a relatively stable level.

\textbf{\emph{iteration:}} In order to find an optimal value for iteration, we train 3 different models to play 6$\times$6 Othello by setting iteration as minimum, default and maximum value respectively and keeping rest parameters as their default values. Overall, either from Fig.~\ref{fig:subfigiteration:a} or Fig.~\ref{fig:subfigiteration:b}, the training loss decreases and elo rating increases both to the relatively stable level. However, after 110th iteration, the training loss unexpectedly increases then deceases again to the same level as only training 100 iteration. This unexpected result could be caused by too big learning rate, improper update threshold, uncertainty from random and so on. Theoretically, we always belief more iteration must lead to better performance, this unexpected result proves the importance of setting proper parameter values.

\begin{figure}[H]
\centering
\hspace*{-2.3em}
\subfigure[Training Loss]{\label{fig:subfigiteration:a} %% label for first subfigure
\includegraphics[width=0.53\textwidth]{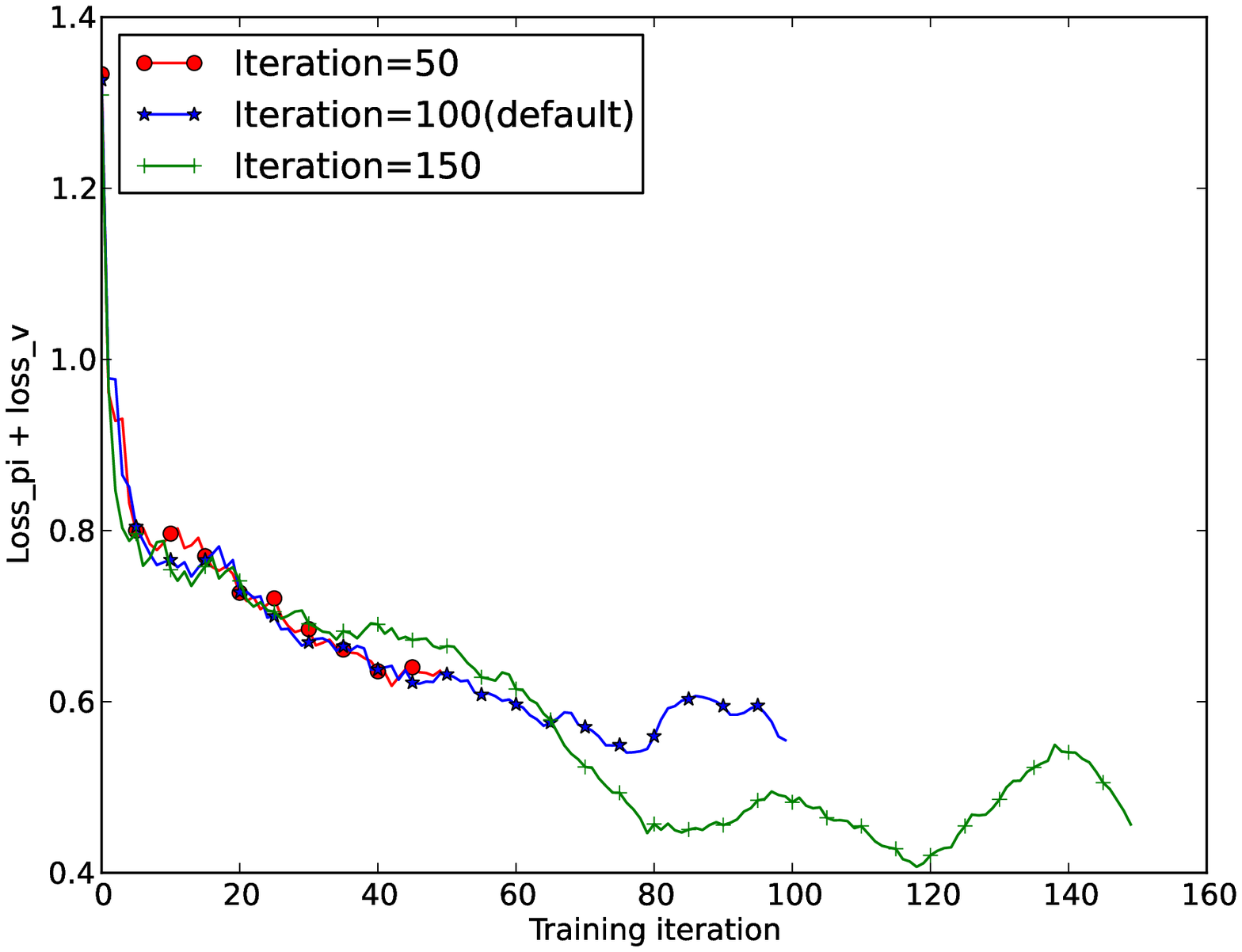}}
\hspace*{-2.3em}
\subfigure[Elo Rating]{\label{fig:subfigiteration:b} %% label for second subfigure
\includegraphics[width=0.53\textwidth]{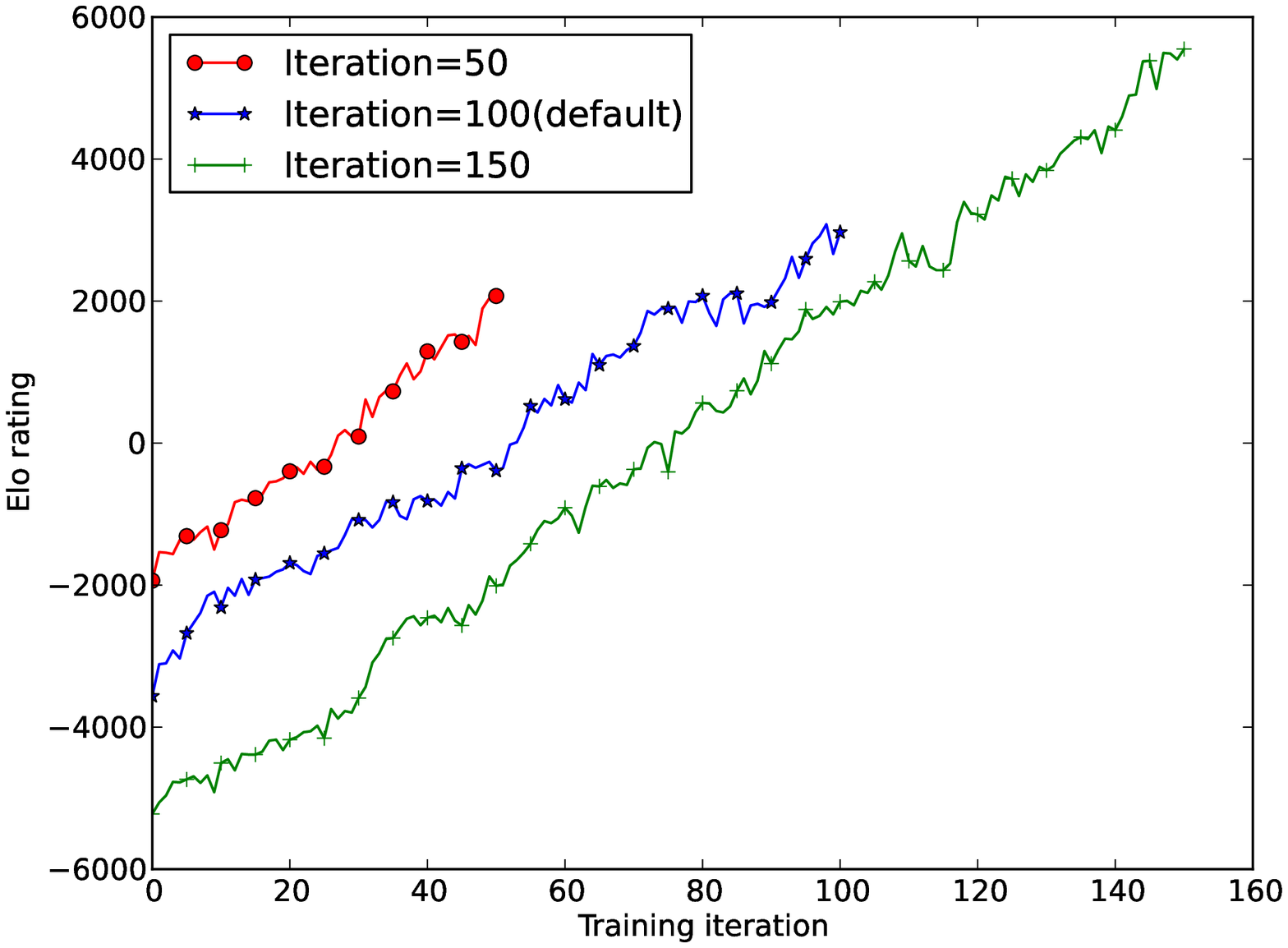}}
%\hspace*{-2.3em}
%\subfigure[Time Cost]{\label{fig:subfigiteration:c} %% label for second subfigure
%\includegraphics[width=0.53\textwidth]{time_iteration.eps}}
\caption{Training Loss and Elo Rating with Different Iteration}
\label{fig:subfigiteration} %% label for entire figure
\end{figure}

\textbf{\emph{episode:}} Since more episode means more training examples. It is believed that more training examples lead to more accurate results. However, collecting more training examples needs more resources. This paradox also suggests that parameter optimization is necessary to find a reasonable value of episode. In Fig.~\ref{fig:subfigepisode:a}, for episode=10, the training loss curve is higher than other 2 curves. However, it is difficult to judge which one is better for other 2 curves, which shows that to much training data does not get significant training results. %From Fig.~\ref{fig:subfigiteration:b}
\begin{figure}[H]
\centering
\hspace*{-2.3em}
\subfigure[Training Loss]{\label{fig:subfigepisode:a} %% label for first subfigure
\includegraphics[width=0.53\textwidth]{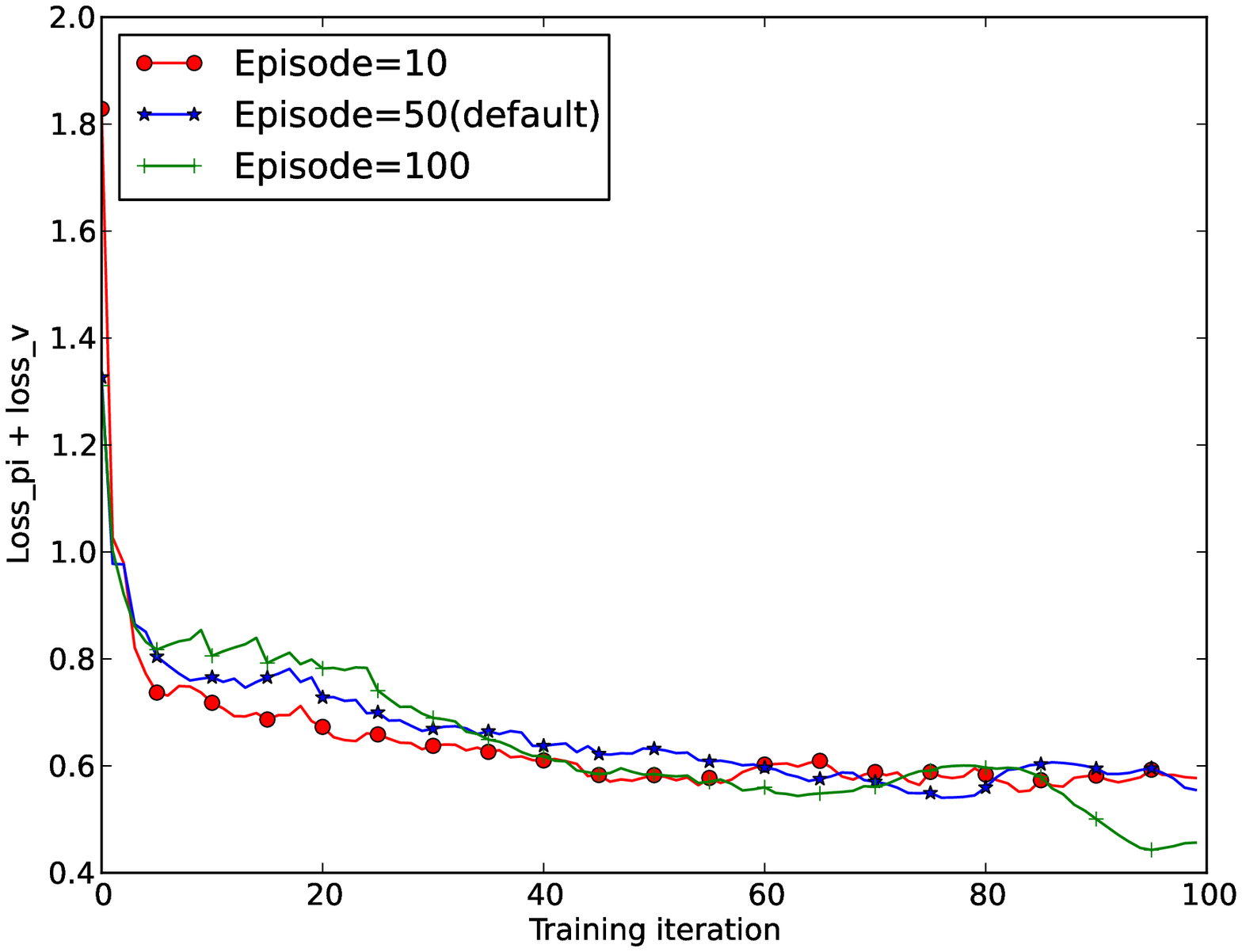}}
\hspace*{-2.3em}
\subfigure[Elo Rating]{\label{fig:subfigepisode:b} %% label for second subfigure
\includegraphics[width=0.53\textwidth]{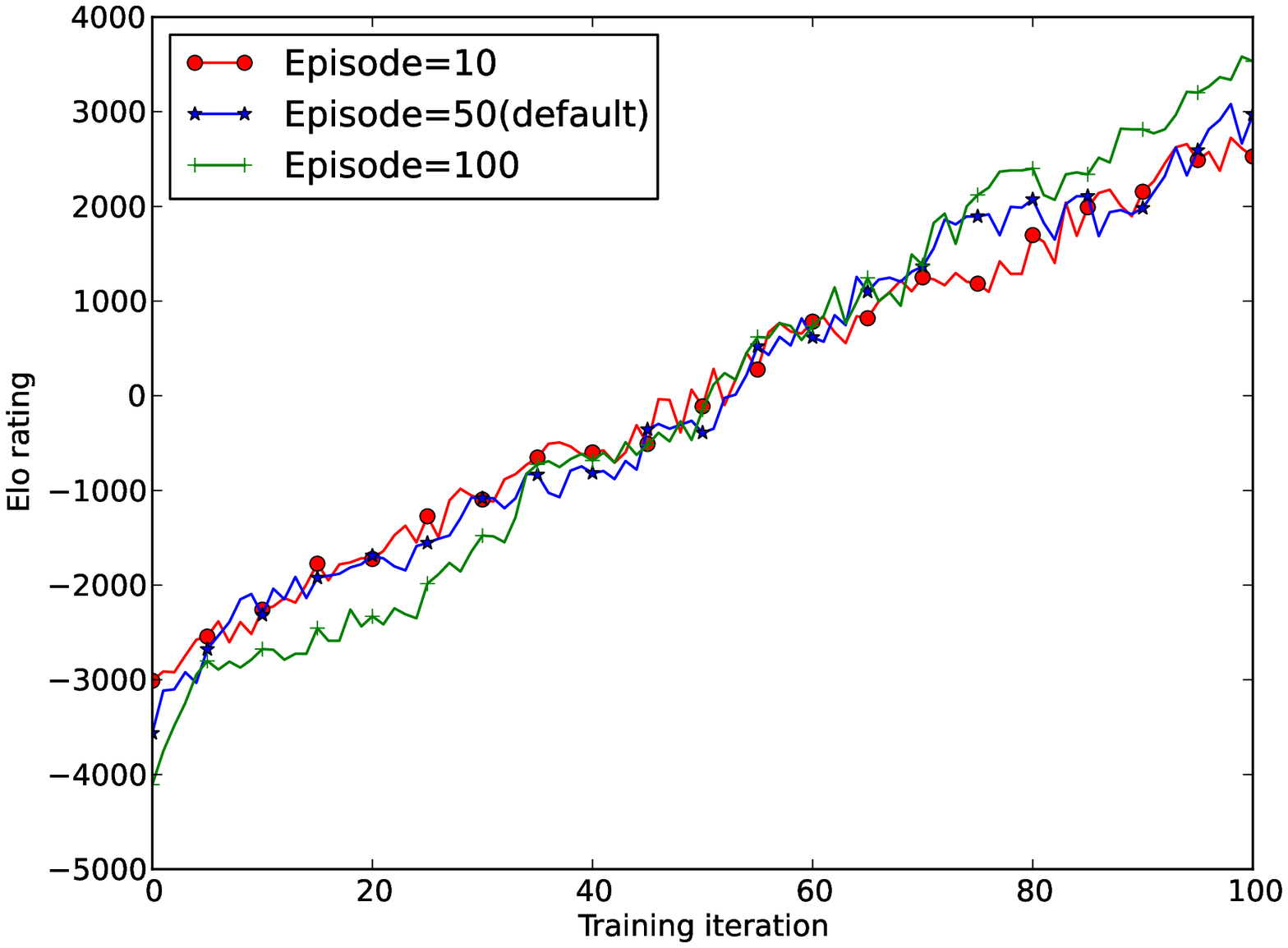}}
%\hspace*{-2.3em}
%\subfigure[Time Cost]{\label{fig:subfigepisode:c} %% label for second subfigure
%\includegraphics[width=0.53\textwidth]{time_episode.eps}}
\caption{Training Loss and Elo Rating with Different Episode}
\label{fig:subfigepisode} %% label for entire figure
\end{figure}

\textbf{\emph{tempThreshold:}} Diversity training examples provide more features, but using too deterministic policy to generate training examples will lead to too many similar~(even the same) training examples in self-play. An option of solving this problem is using tempThreshold to control it. However, too small tempThreshold makes policy more deterministic, too big tempThreshold makes policy more different from the model. From Fig.~\ref{fig:subfigtempThreshold:a} and Fig.~\ref{fig:subfigtempThreshold:b}, we see tempThreshold=10 is the best value.
\begin{figure}[H]
\centering
\hspace*{-2.3em}
\subfigure[Training Loss]{\label{fig:subfigtempThreshold:a} %% label for first subfigure
\includegraphics[width=0.53\textwidth]{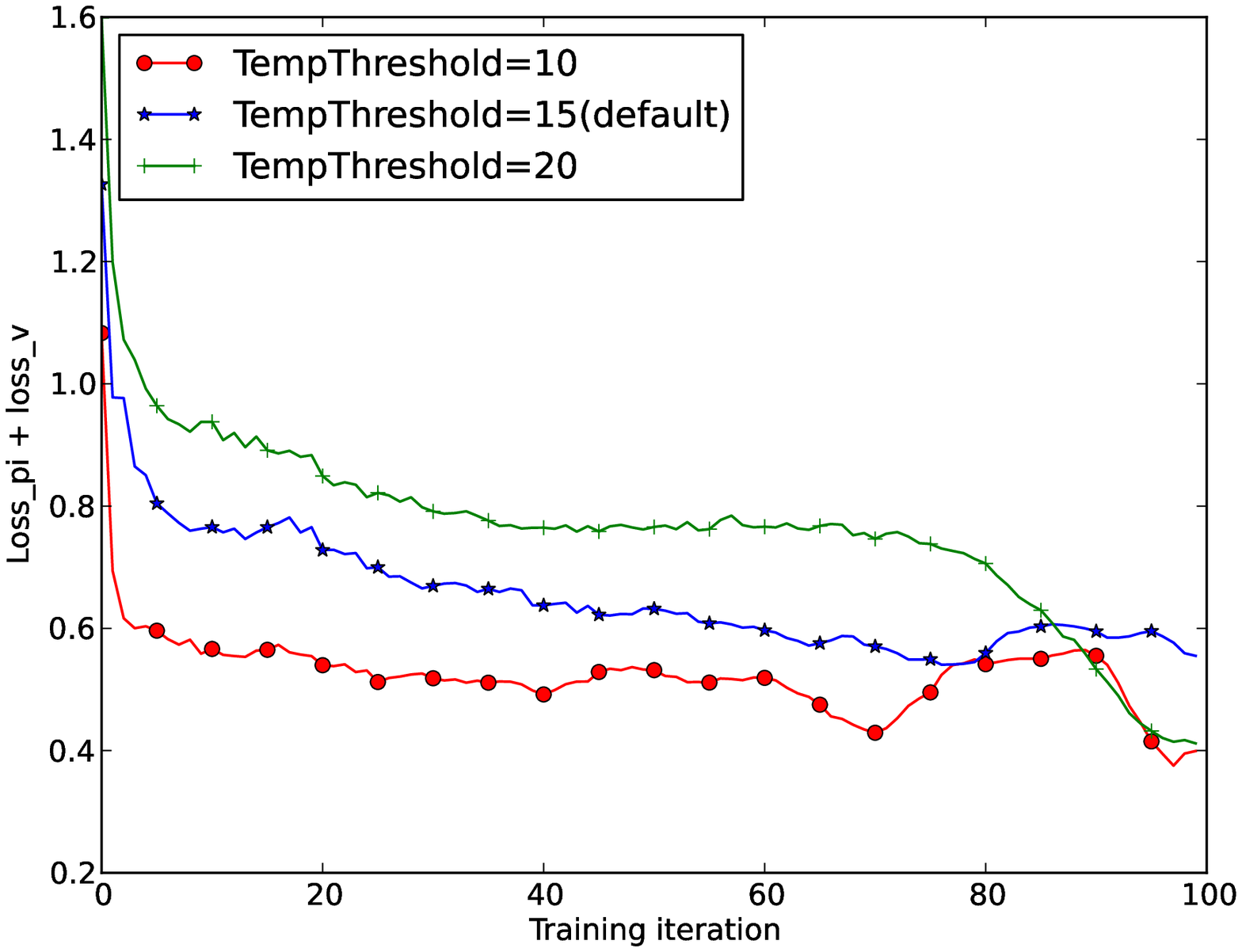}}
\hspace*{-2.3em}
\subfigure[Elo Rating]{\label{fig:subfigtempThreshold:b} %% label for second subfigure
\includegraphics[width=0.53\textwidth]{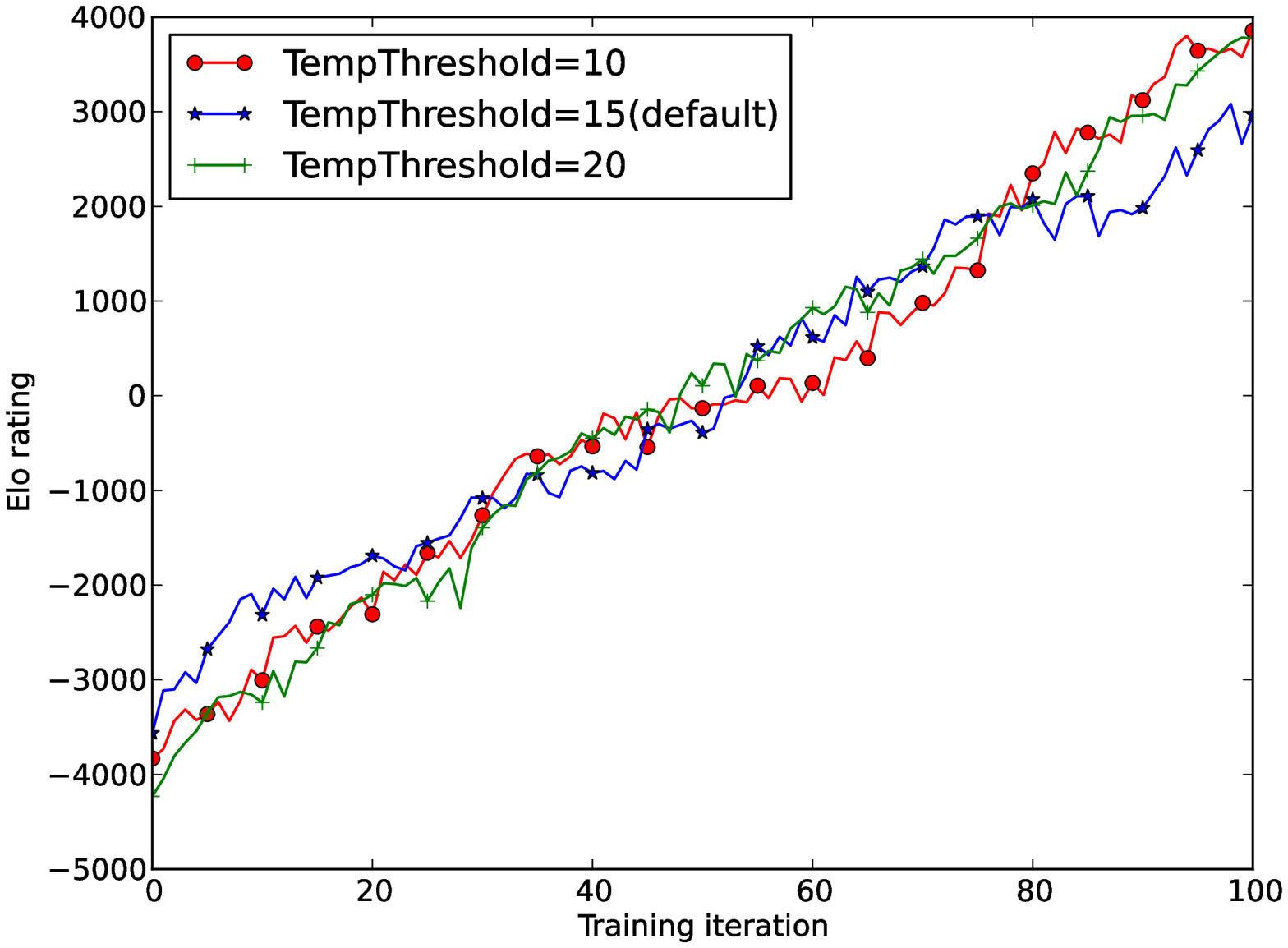}}
%\hspace*{-2.3em}
%\subfigure[Time Cost]{\label{fig:subfigtempThreshold:c} %% label for second subfigure
%\includegraphics[width=0.53\textwidth]{time_tempThreshold.eps}}
\caption{Training Loss and Elo Rating with Different TempThreshold}
\label{fig:subfigtempThreshold} %% label for entire figure
\end{figure}

\textbf{\emph{mctssimu:}} In theory, more mctssimu can provide better policy for performing the best move. However, more mctssimu means more time to get such good policy. Besides, In Algorithm~\ref{alg:a0g}, tree search is also used to generate an estimation improvement $\overrightarrow{\pi}$ to calculate the cross entropy for loss function. In Fig.~\ref{fig:subfigmctssimu:a} and Fig.~\ref{fig:subfigmctssimu:b}, 200 times simulation achieves best performance in 60th iteration then unexpectedly performs worse and reaches the same level as 100 simulation times in 100th iteration.
\begin{figure}[H]
\centering
\hspace*{-2.3em}
\subfigure[Training Loss]{\label{fig:subfigmctssimu:a} %% label for first subfigure
\includegraphics[width=0.53\textwidth]{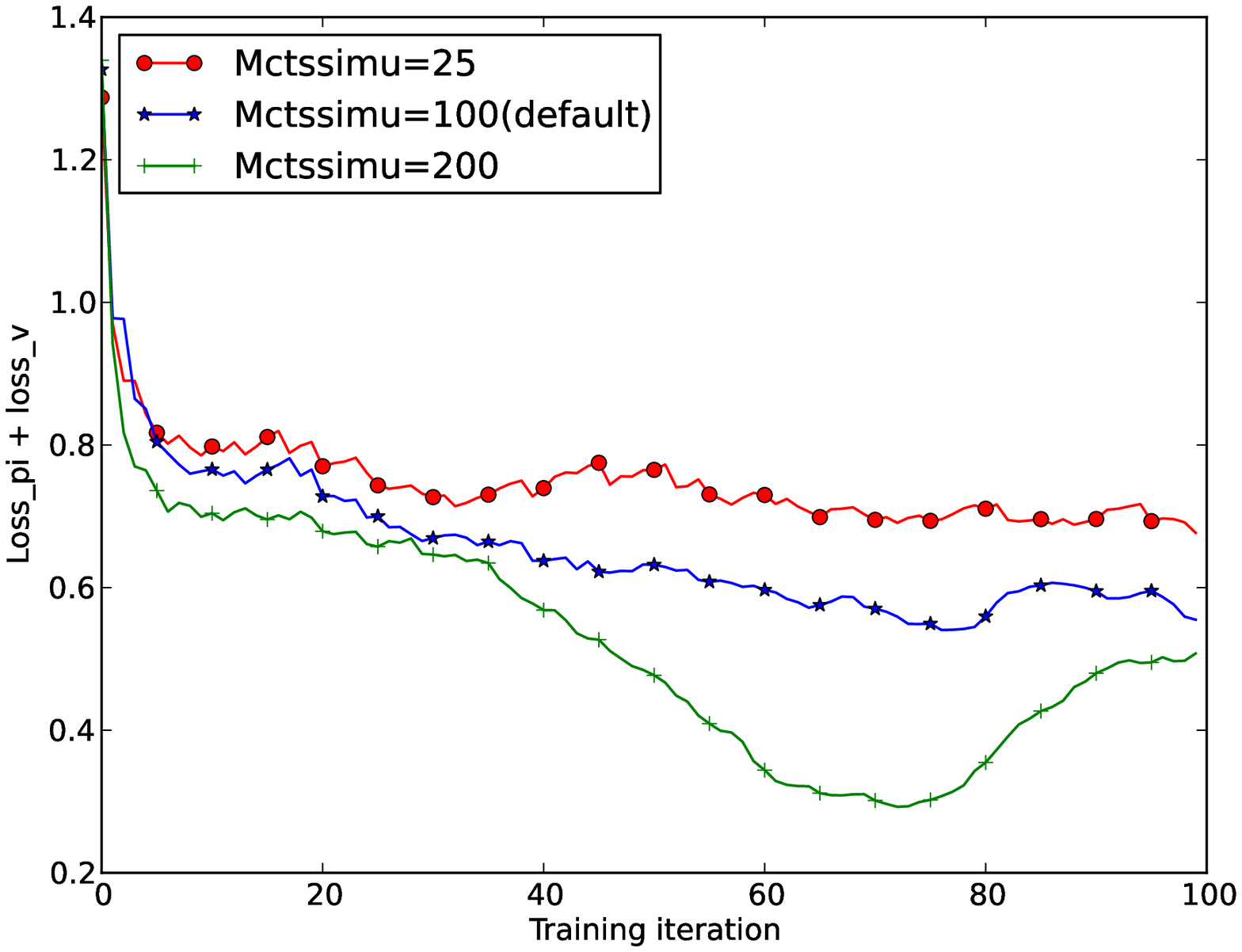}}
\hspace*{-2.3em}
\subfigure[Elo Rating]{\label{fig:subfigmctssimu:b} %% label for second subfigure
\includegraphics[width=0.53\textwidth]{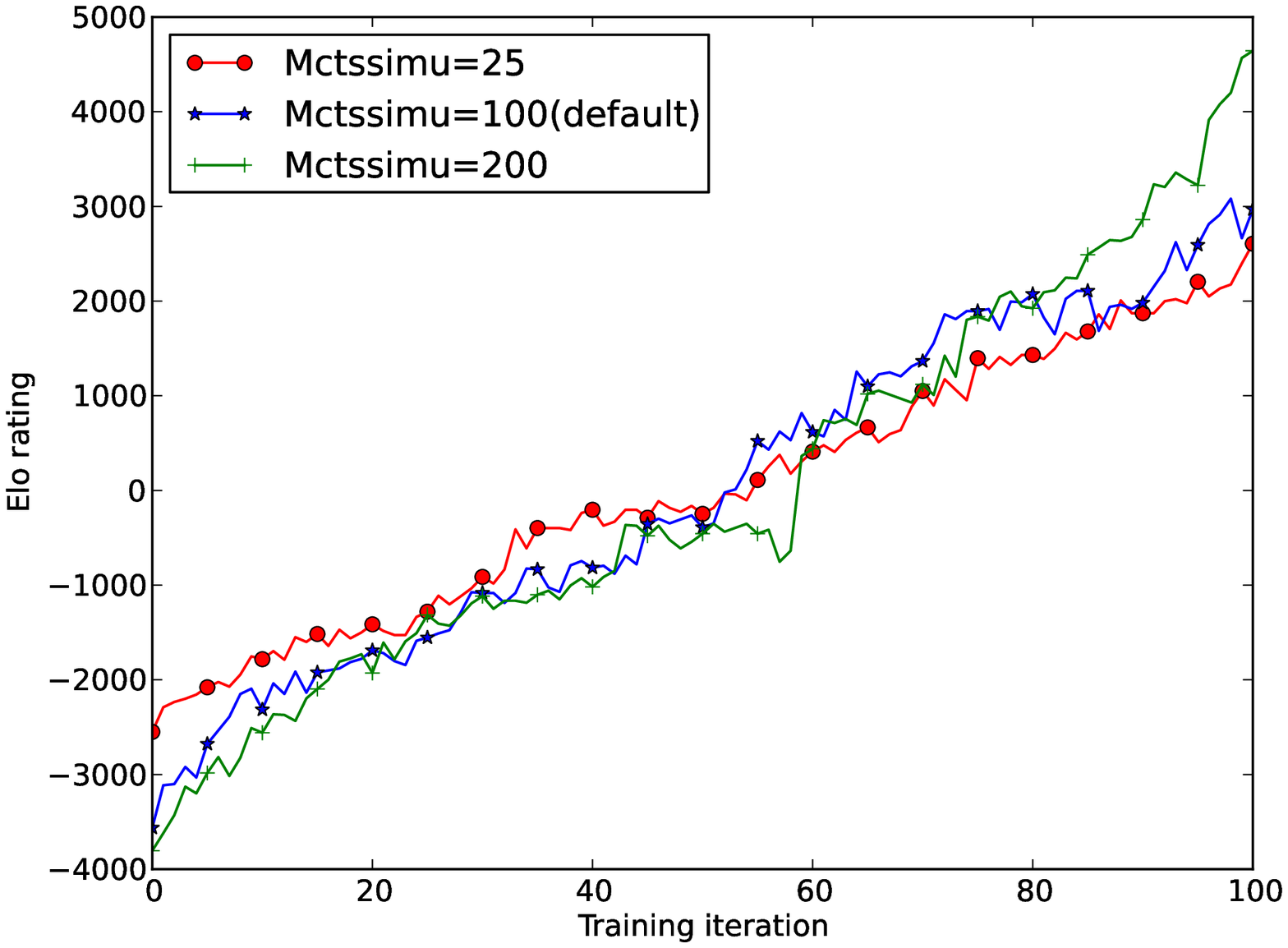}}
%\hspace*{-2.3em}
%\subfigure[Time Cost]{\label{fig:subfigmctssimu:c} %% label for second subfigure
%\includegraphics[width=0.53\textwidth]{time_mctssimu.eps}}
\caption{Training Loss and Elo Rating with Different Mctssimu}
\label{fig:subfigmctssimu} %% label for entire figure
\end{figure}

\textbf{\emph{Cpuct:}} This parameter is used to balance the exploration and exploitation during tree search. Practically, it is suggested to set as 1.0. However, in Fig.~\ref{fig:subfigCpuct:a}, our experimental results show that more exploitation can get smaller training loss. %In n Fig.~\ref{fig:subfigCpuct:b}
\begin{figure}[H]
\centering
\hspace*{-2.3em}
\subfigure[Training Loss]{\label{fig:subfigCpuct:a} %% label for first subfigure
\includegraphics[width=0.53\textwidth]{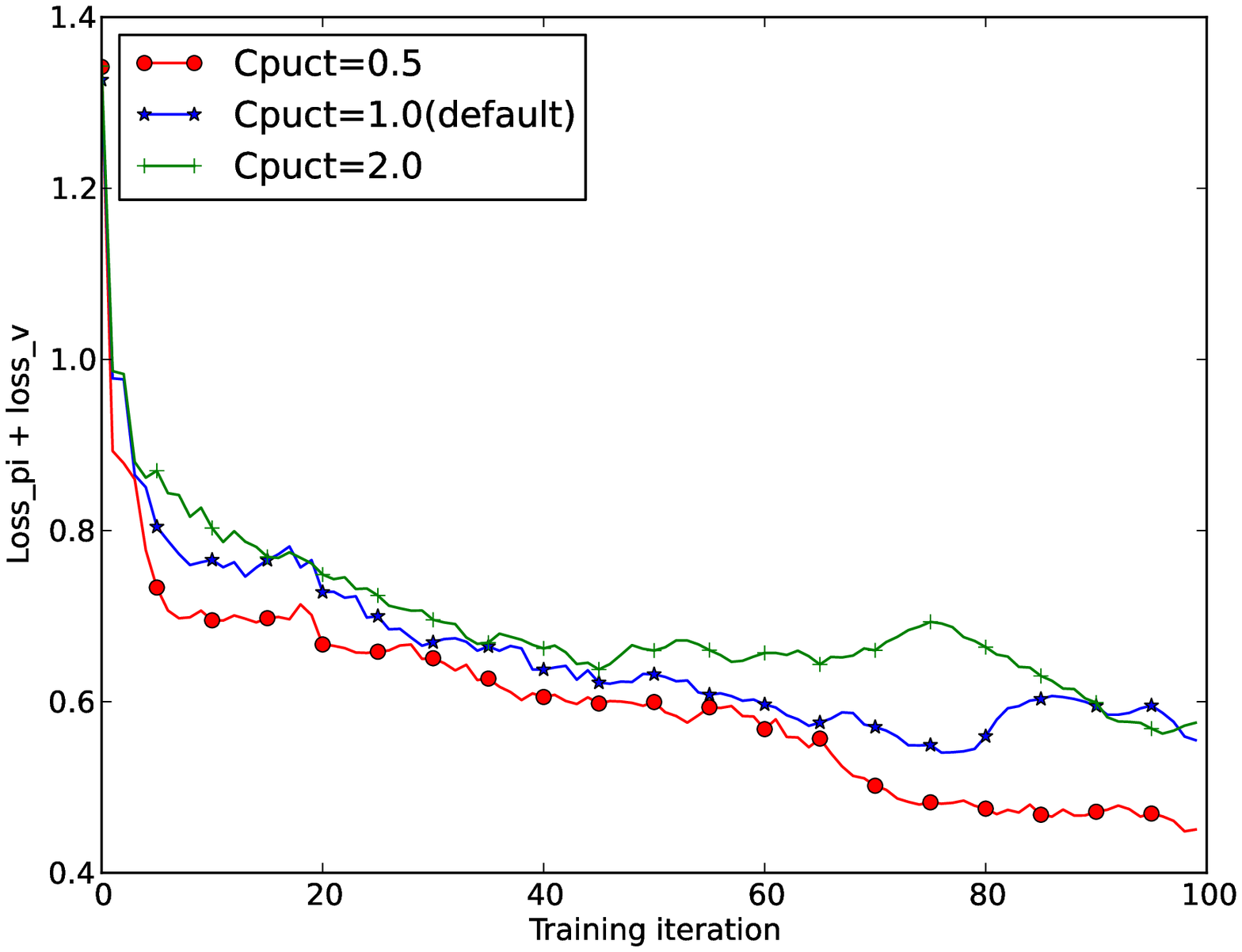}}
\hspace*{-2.3em}
\subfigure[Elo Rating]{\label{fig:subfigCpuct:b} %% label for second subfigure
\includegraphics[width=0.53\textwidth]{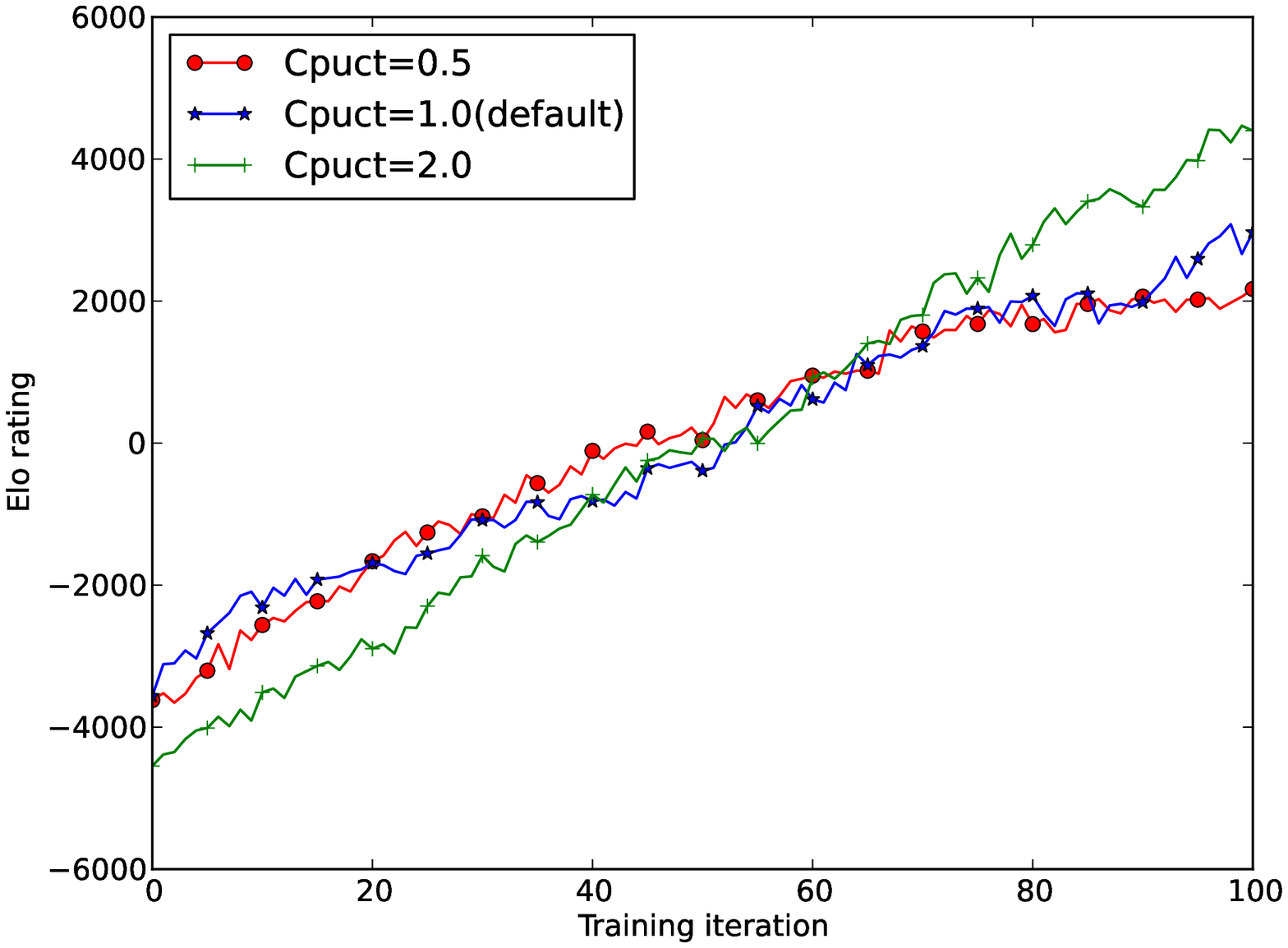}}
%\hspace*{-2.3em}
%\subfigure[Time Cost]{\label{fig:subfigCpuct:c} %% label for second subfigure
%\includegraphics[width=0.53\textwidth]{time_Cpuct.eps}}
\caption{Training Loss and Elo Rating with Different Cpuct}
\label{fig:subfigCpuct} %% label for entire figure
\end{figure}

\textbf{\emph{retrainlength:}} In order to reduce overfitting, it is important to retrain model using previous training examples. Finding an optimal retrain length of historical training examples is necessary to reduce waste of time. In Fig.~\ref{fig:subfigretrain:a}, we see that, only using training examples from most recent 1 iteration achieves smallest training loss, which seems unexpected. However, in Fig.~\ref{fig:subfigretrain:b}, we find that using training examples from previous 40 iterations gets a stable elo rating quickly~(in 60th iteration). In fact, less retrain examples leads to overfitting so that an unexpected result shows up.
\begin{figure}[H]
\centering
\hspace*{-2.3em}
\subfigure[Training Loss]{\label{fig:subfigretrain:a} %% label for first subfigure
\includegraphics[width=0.53\textwidth]{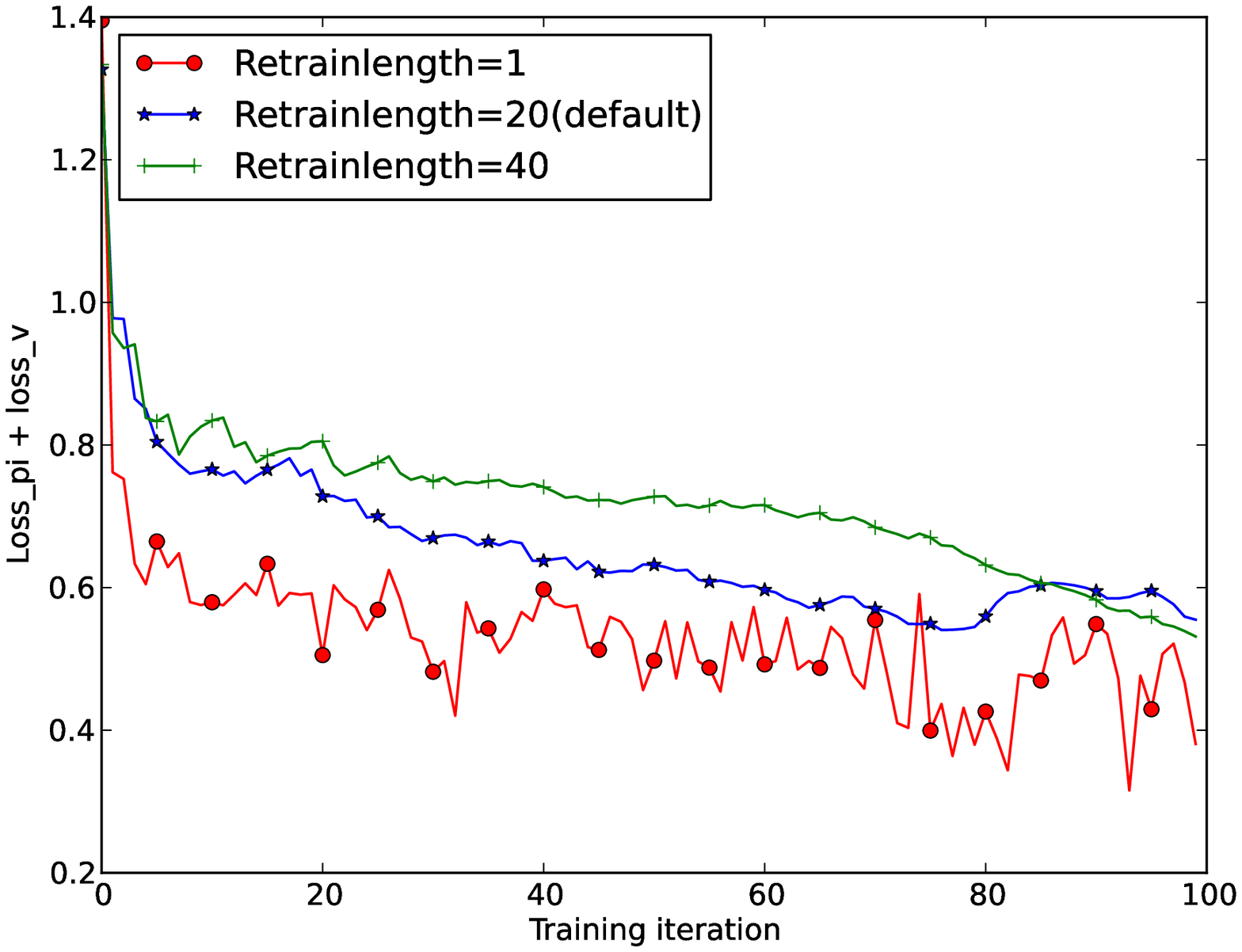}}
\hspace*{-2.3em}
\subfigure[Elo Rating]{\label{fig:subfigretrain:b} %% label for second subfigure
\includegraphics[width=0.53\textwidth]{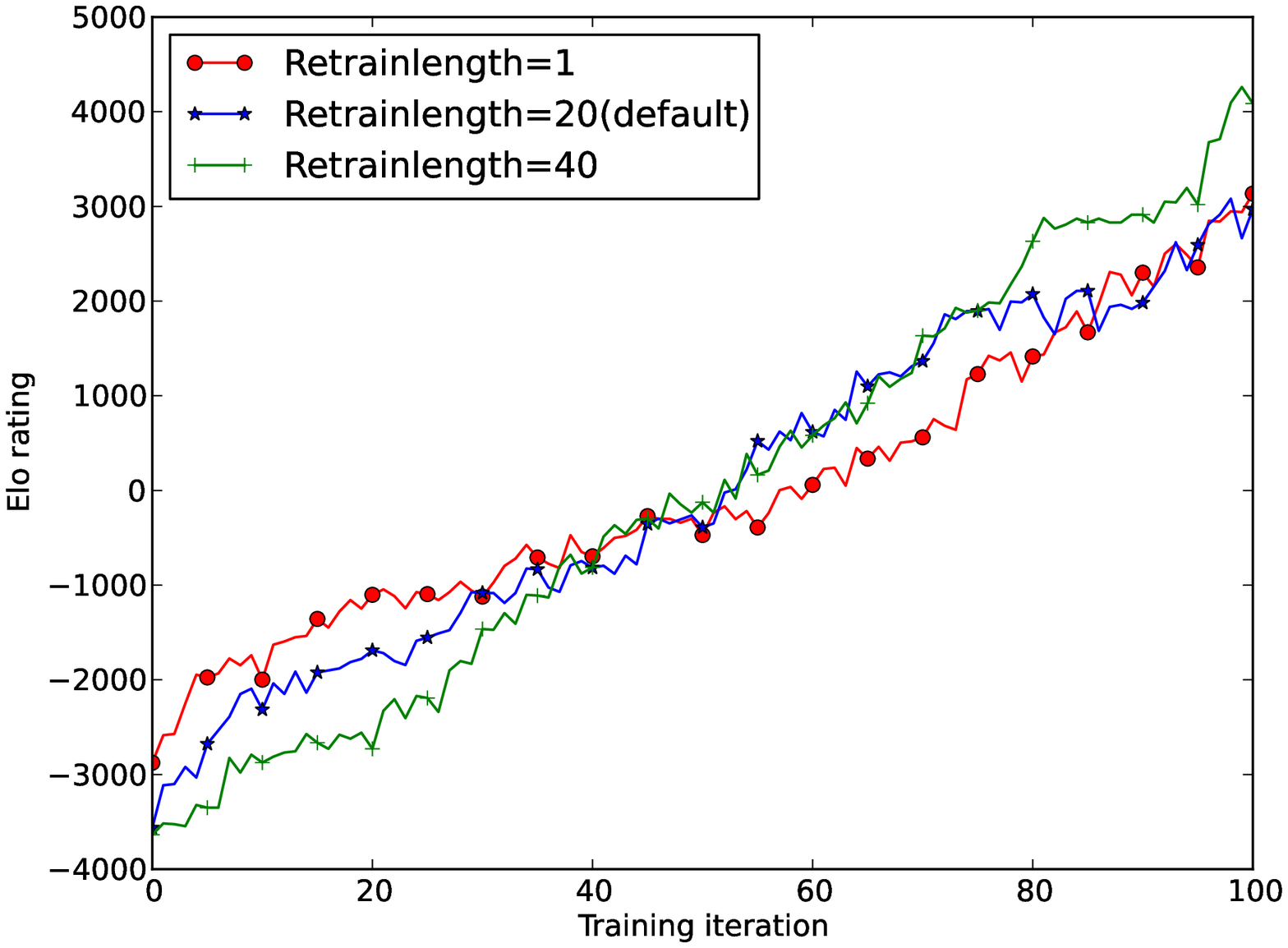}}
%\hspace*{-2.3em}
%\subfigure[Time Cost]{\label{fig:subfigretrain:c} %% label for second subfigure
%\includegraphics[width=0.53\textwidth]{time_retrain.eps}}
\caption{Training Loss and Elo Rating with Different Retrainlength}
\label{fig:subfigretrain} %% label for entire figure
\end{figure}

\textbf{\emph{epoch:}} The training loss of different epoch is presented as Fig.~\ref{fig:subfigepoch:a}. As we can see, while epoch=15, the training loss curve is the lowest. The smallest training loss in the 90th iteration, which is about 0.2. The experimental results prove that along with the increasing of epoch, the training loss decreases, which is in expected.
\begin{figure}[H]
\centering
\hspace*{-2.3em}
\subfigure[Training Loss]{\label{fig:subfigepoch:a} %% label for first subfigure
\includegraphics[width=0.53\textwidth]{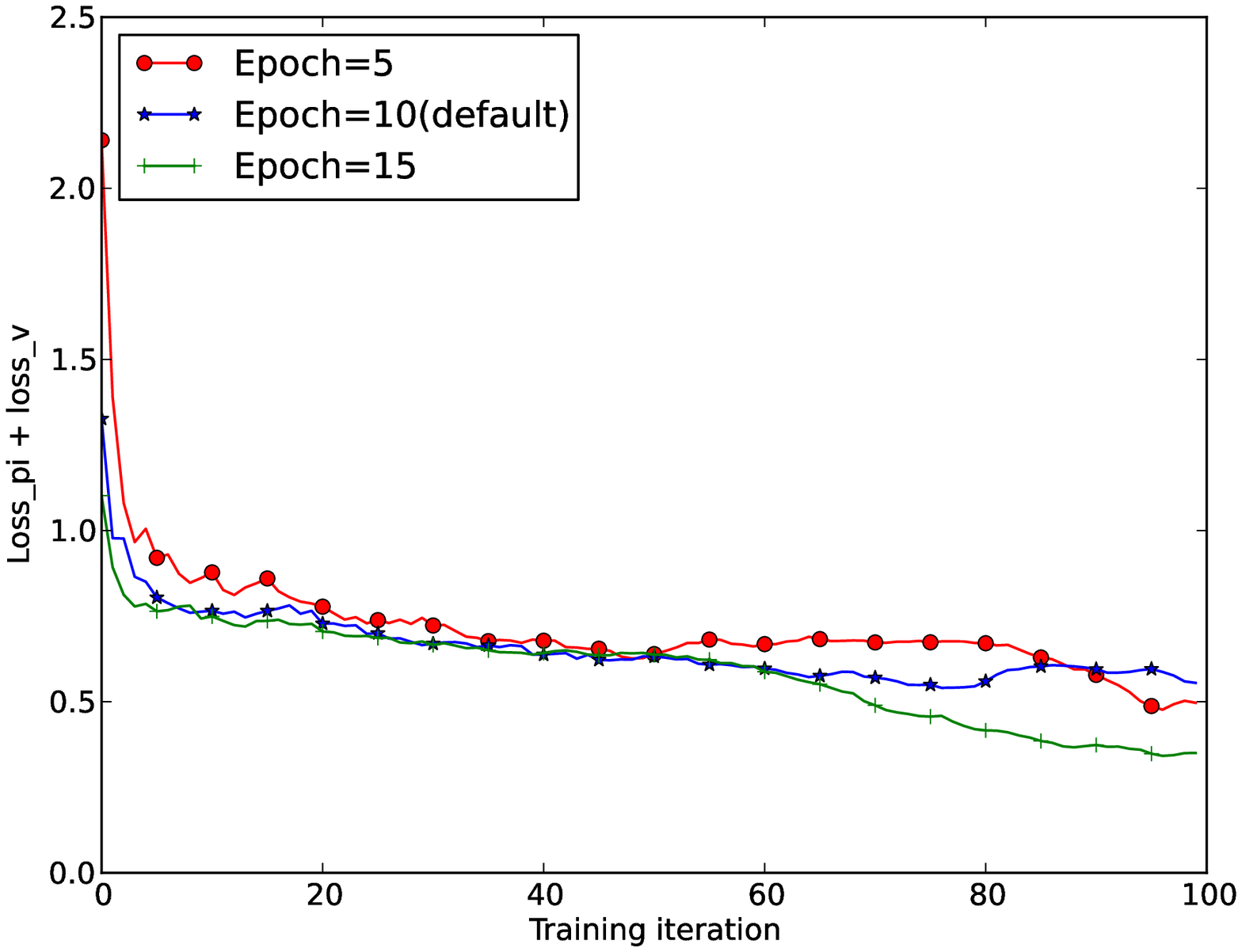}}
\hspace*{-2.3em}
\subfigure[Elo Rating]{\label{fig:subfigepoch:b} %% label for second subfigure
\includegraphics[width=0.53\textwidth]{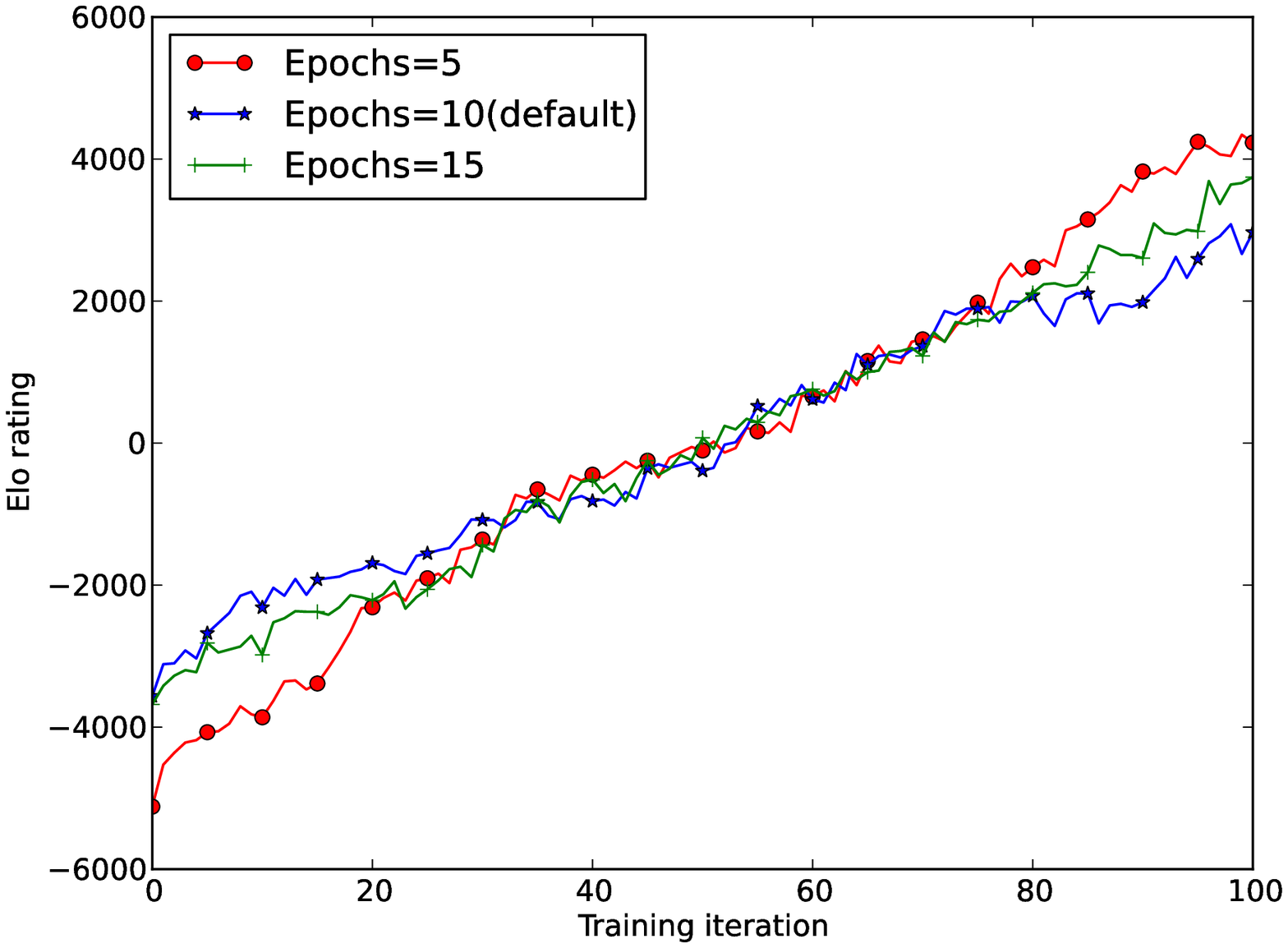}}
%\hspace*{-2.3em}
%\subfigure[Time Cost]{\label{fig:subfigepoch:c} %% label for second subfigure
%\includegraphics[width=0.53\textwidth]{time_epoch.eps}}
\caption{Training Loss and Elo Rating with Different Epoch}
\label{fig:subfigepoch} %% label for entire figure
\end{figure}

\textbf{\emph{batchsize:}} On the one hand, smaller batchsize makes the number of batches become larger, which will lead to more time cost. However, on the other hand, smaller batchsize means less training examples in each batch, which could cause more fluctuation~(larger variance) of training loss. From Fig.~\ref{fig:subfigbatchsize:a}, batchsize=32 achieves the smallest training loss in 94th iteration. And in Fig.~\ref{fig:subfigbatchsize:b}, batchsize=32 reaches the highest stable elo rating more quick than others.
\begin{figure}[H]
\centering
\hspace*{-2.3em}
\subfigure[Training Loss]{\label{fig:subfigbatchsize:a} %% label for first subfigure
\includegraphics[width=0.53\textwidth]{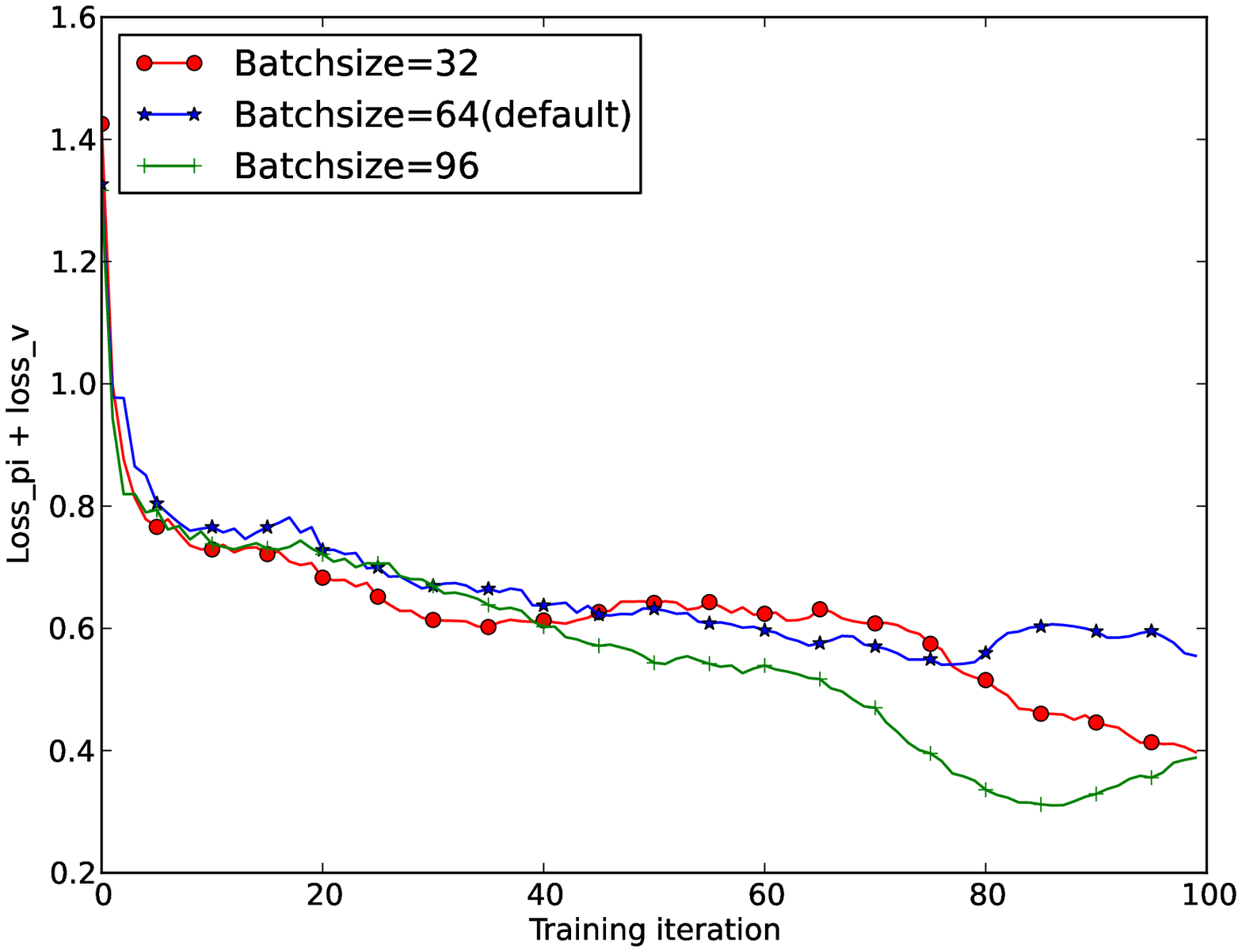}}
\hspace*{-2.3em}
\subfigure[Elo Rating]{\label{fig:subfigbatchsize:b} %% label for second subfigure
\includegraphics[width=0.53\textwidth]{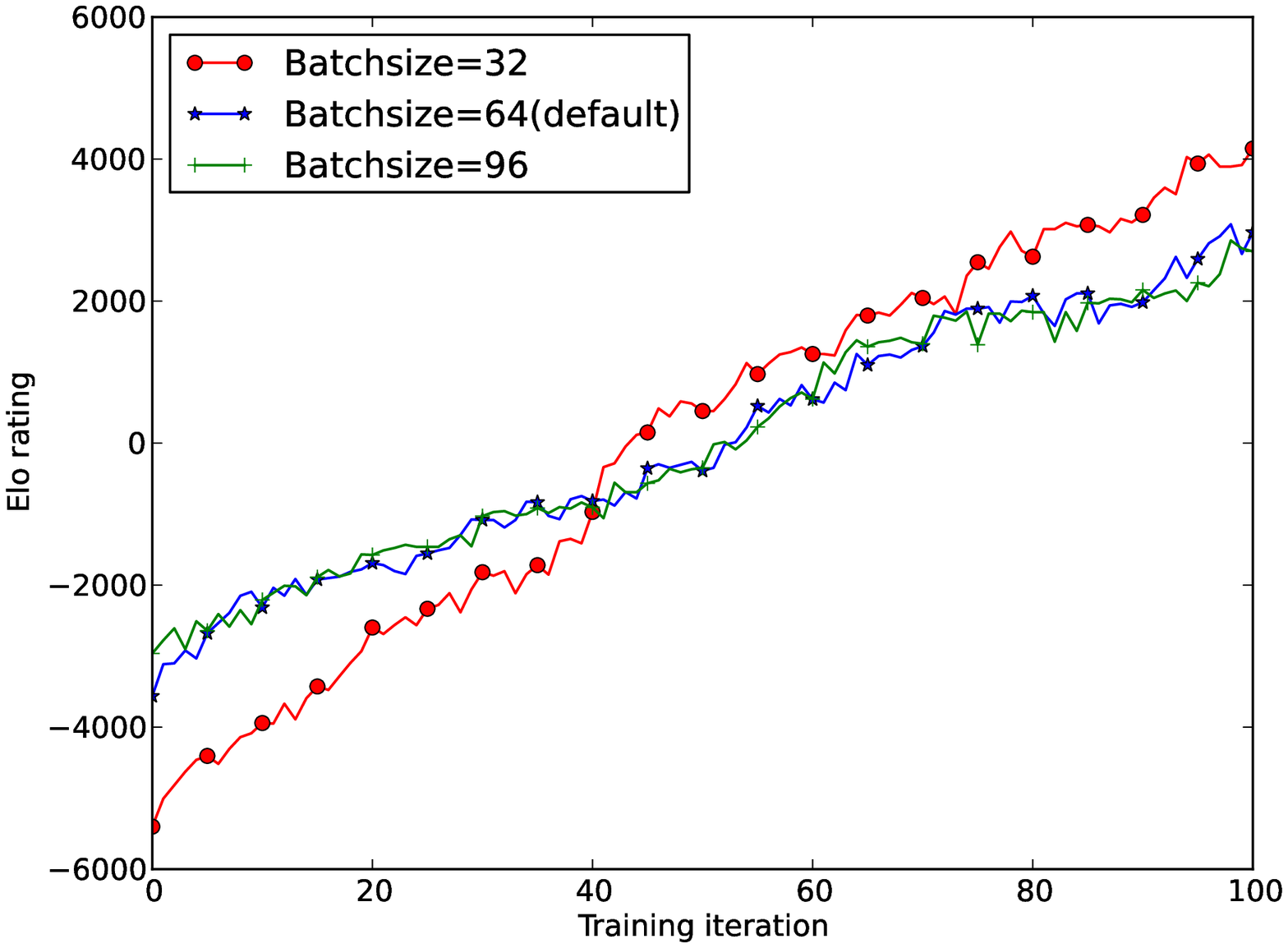}}
%\hspace*{-2.3em}
%\subfigure[Time Cost]{\label{fig:subfigbatchsize:c} %% label for second subfigure
%\includegraphics[width=0.53\textwidth]{time_batchsize.eps}}
\caption{Training Loss and Elo Rating with Different Batchsize}
\label{fig:subfigbatchsize} %% label for entire figure
\end{figure}

\textbf{\emph{learningrate:}} In order to avoid falling into a local optimal, normally, a smaller learning rate is suggested. However, a smaller learning rate learns~(accepts) new knowledge slowly. In Fig.~\ref{fig:subfiglearningrate:a}, learningrate=0.001 achieves the lowest training loss which is in 93th iteration. From Fig.~\ref{fig:subfiglearningrate:b}, while learningrate=0.01, the elo rating is in expected to get higher quickly but falls down at last unluckily.
\begin{figure}[H]
\centering
\hspace*{-2.3em}
\subfigure[Training Loss]{\label{fig:subfiglearningrate:a} %% label for first subfigure
\includegraphics[width=0.53\textwidth]{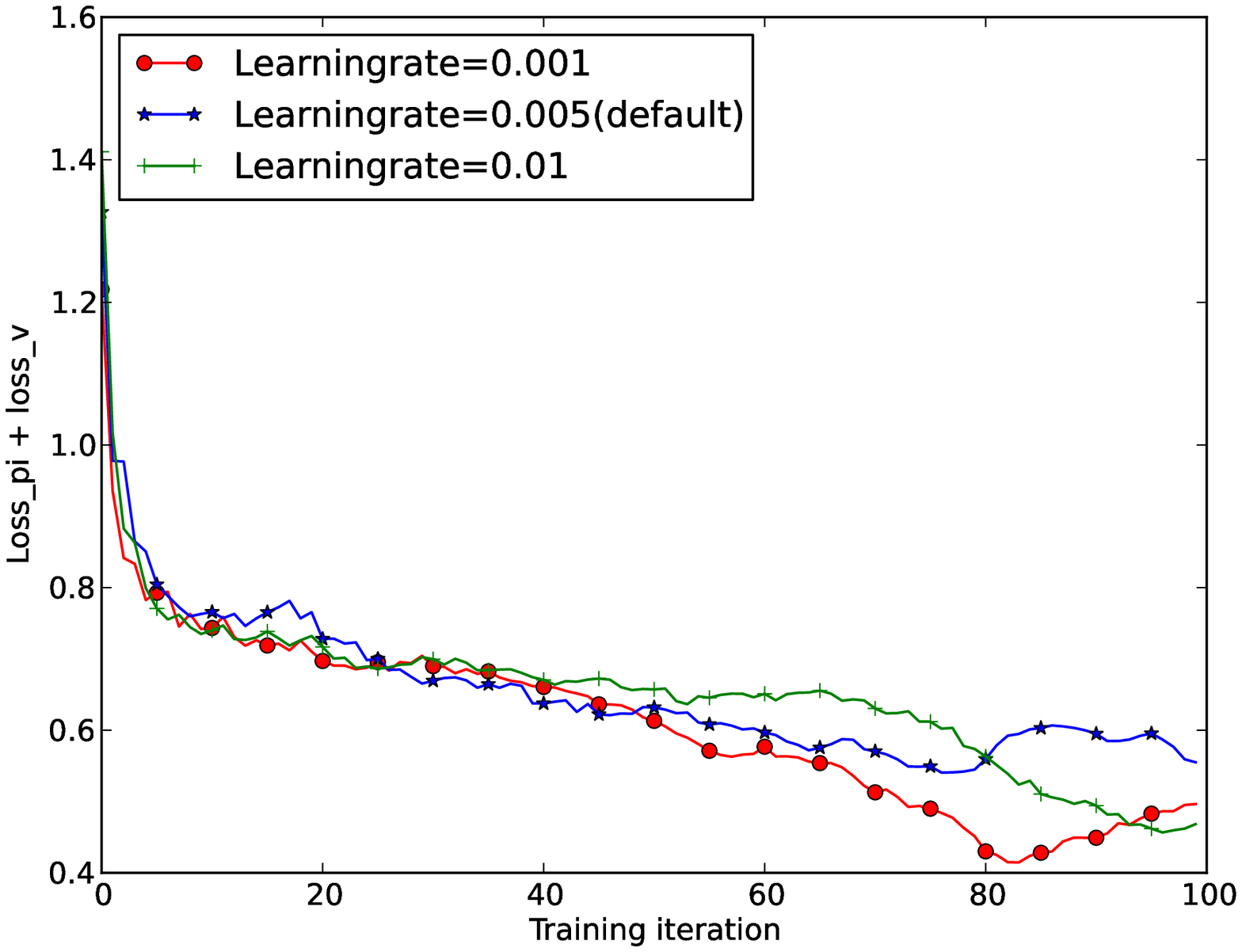}}
\hspace*{-2.3em}
\subfigure[Elo Rating]{\label{fig:subfiglearningrate:b} %% label for second subfigure
\includegraphics[width=0.53\textwidth]{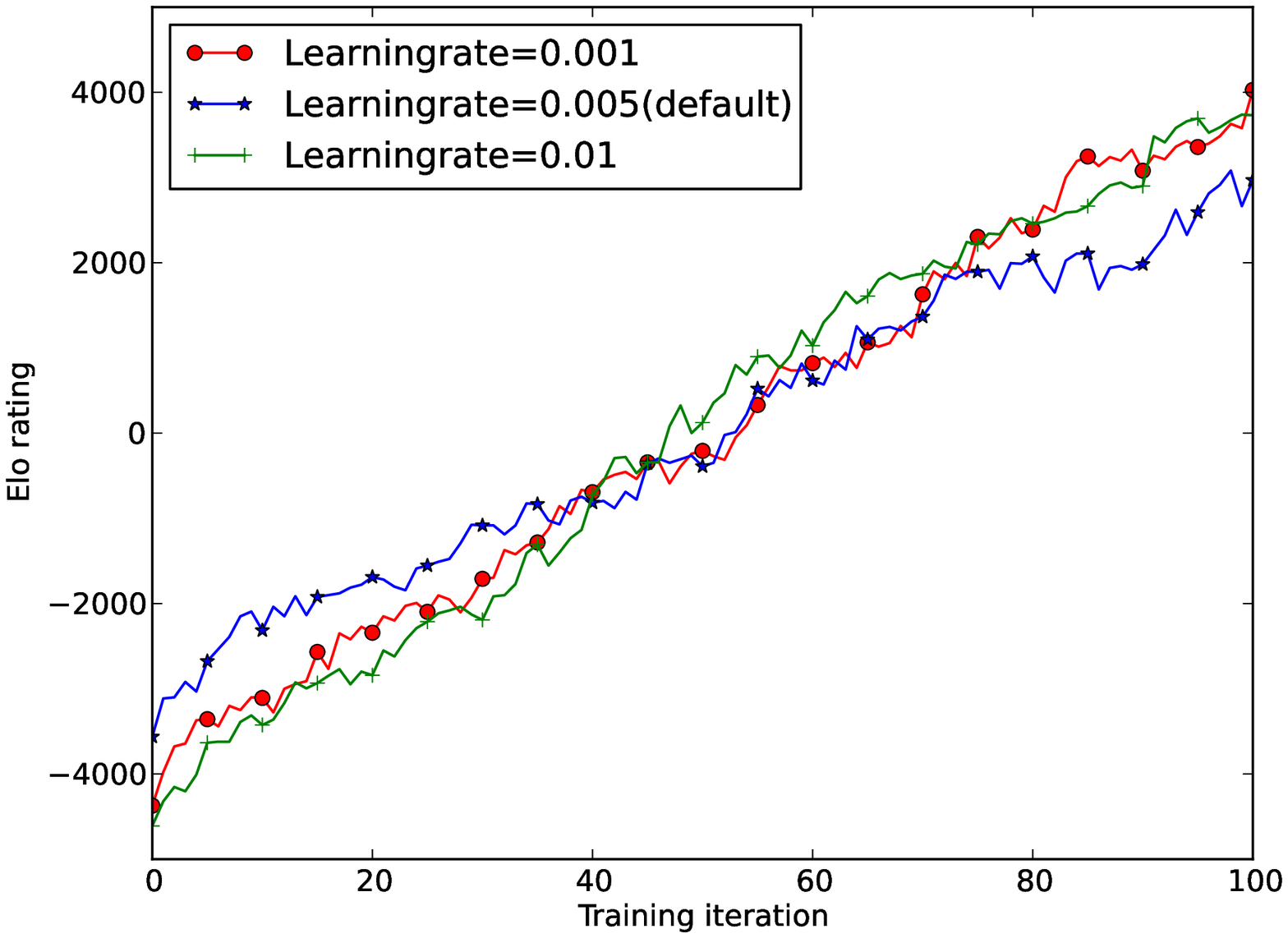}}
%\hspace*{-2.3em}
%\subfigure[Time Cost]{\label{fig:subfiglearningrate:c} %% label for second subfigure
%\includegraphics[width=0.53\textwidth]{time_learningrate.eps}}
\caption{Training Loss and Elo Rating with Different Learningrate}
\label{fig:subfiglearningrate} %% label for entire figure
\end{figure}

\textbf{\emph{dropout:}} Dropout prevents overfitting and provides a way of approximately combining exponentially many different neural network architectures efficiently. Srivastava et al. claims that dropping out 20\% of the input units and 50\% of the hidden units is often found to be optimal~\cite{Srivastava2014}. In Fig.~\ref{fig:subfiglearningrate:a}, we find dropout=0.3 achieves the lowest training loss in 91th iteration, which is about 0.26. %However, from Fig.~\ref{fig:subfiglearningrate:b}
\begin{figure}[H]
\centering
\hspace*{-2.3em}
\subfigure[Training Loss]{\label{fig:subfigdropout:a} %% label for first subfigure
\includegraphics[width=0.53\textwidth]{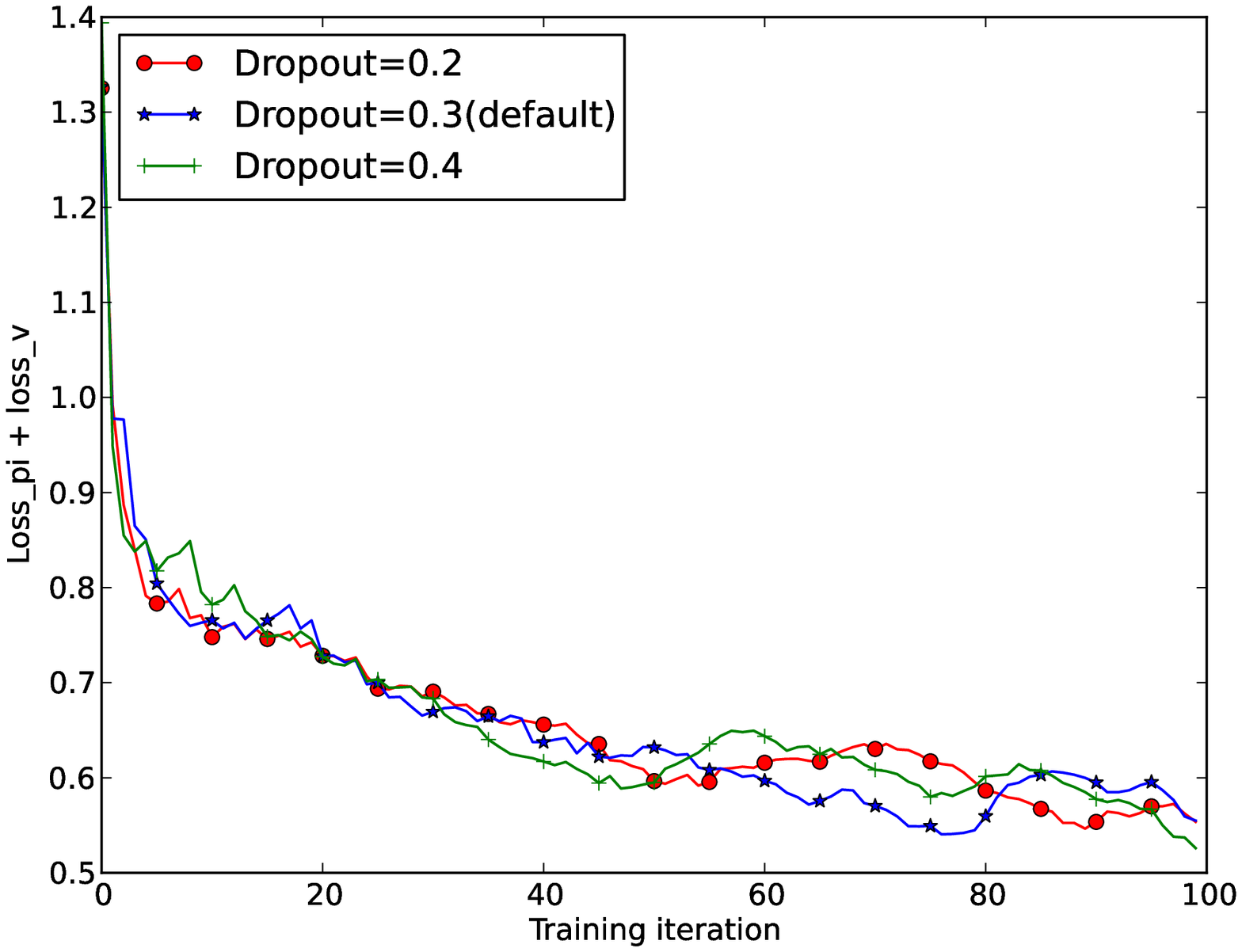}}
\hspace*{-2.3em}
\subfigure[Elo Rating]{\label{fig:subfigdropout:b} %% label for second subfigure
\includegraphics[width=0.53\textwidth]{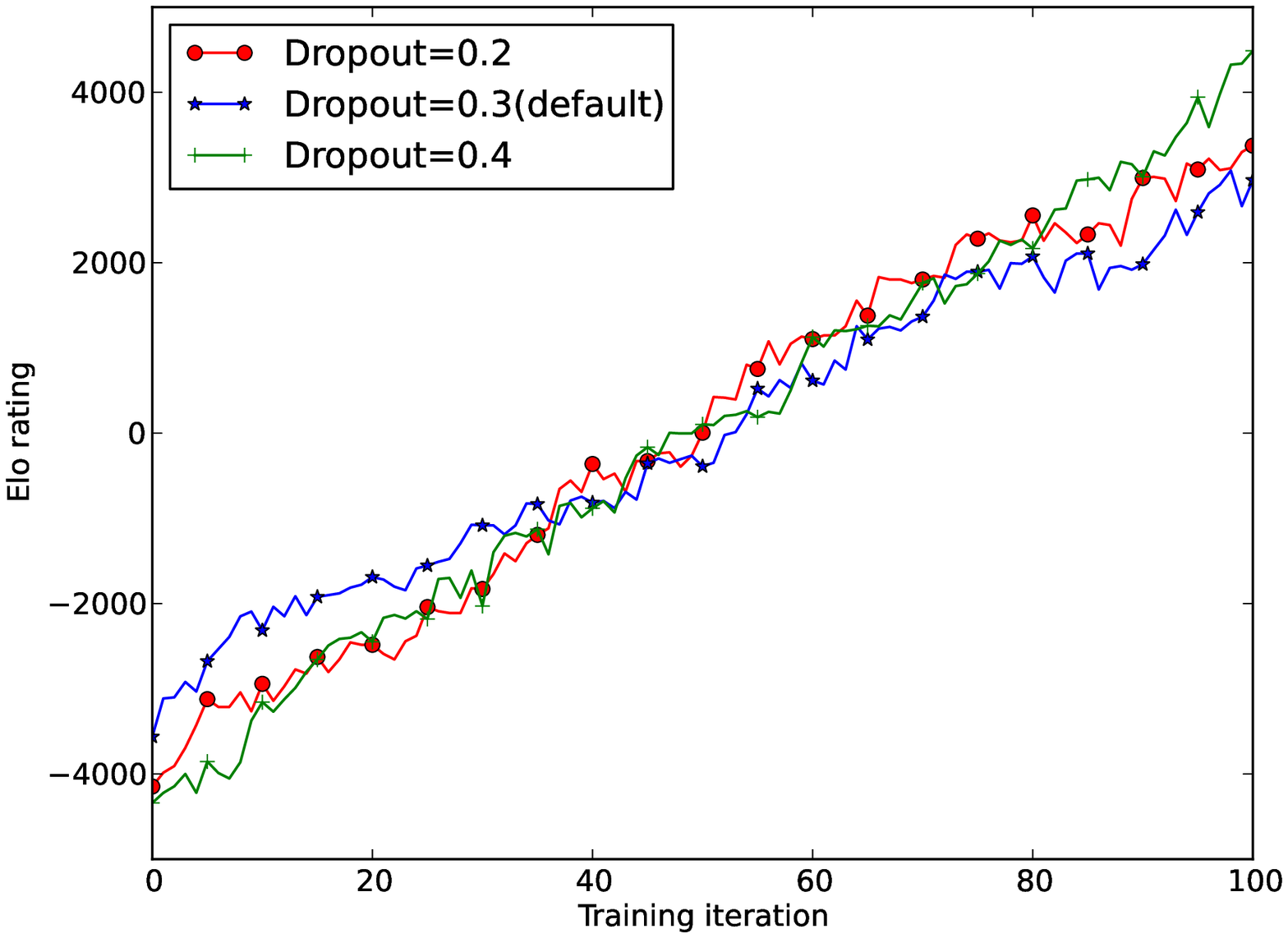}}
%\hspace*{-2.3em}
%\subfigure[Time Cost]{\label{fig:subfigdropout:c} %% label for second subfigure
%\includegraphics[width=0.53\textwidth]{time_dropout.eps}}
\caption{Training Loss and Elo Rating with Different Dropout}
\label{fig:subfigdropout} %% label for entire figure
\end{figure}

%\textbf{\emph{channel:}}
%\begin{figure}[H]
%\centering\hspace*{-2.3em}
%\subfigure[Training Loss]{\label{fig:subfigchannel:a} %% label for first subfigure
%\includegraphics[width=0.53\textwidth]{loss_pi_v_channel.eps}}
%\hspace*{-2.3em}
%\subfigure[Elo Rating]{\label{fig:subfigchannel:b} %% label for second subfigure
%\includegraphics[width=0.53\textwidth]{elo_channel.eps}}
%%\hspace*{-2.3em}
%%\subfigure[Time Cost]{\label{fig:subfigchannel:c} %% label for second subfigure
%%\includegraphics[width=0.53\textwidth]{time_channel.eps}}
%\caption{Training Loss and Elo Rating with Different Channel}
%\label{fig:subfigchannel} %% label for entire figure
%\end{figure}

\textbf{\emph{arenacompare:}} Obviously, this parameter is a key factor of time cost in arena comparison. Too small value can not avoid coincidence and too large value is time-sensitive and not necessary at all. Our experimental results in Fig.~\ref{fig:subfigarenacompare:a} show that there is no significant difference. A combination with updateThreshold can be used to determine the acceptance or rejection of newly learnt model. In order to reduce time cost, a little small arenacompare combines with a litter big updateThreshold might be the proper choice.
\begin{figure}[H]
\centering
\hspace*{-2.3em}
\subfigure[Training Loss]{\label{fig:subfigarenacompare:a} %% label for first subfigure
\includegraphics[width=0.53\textwidth]{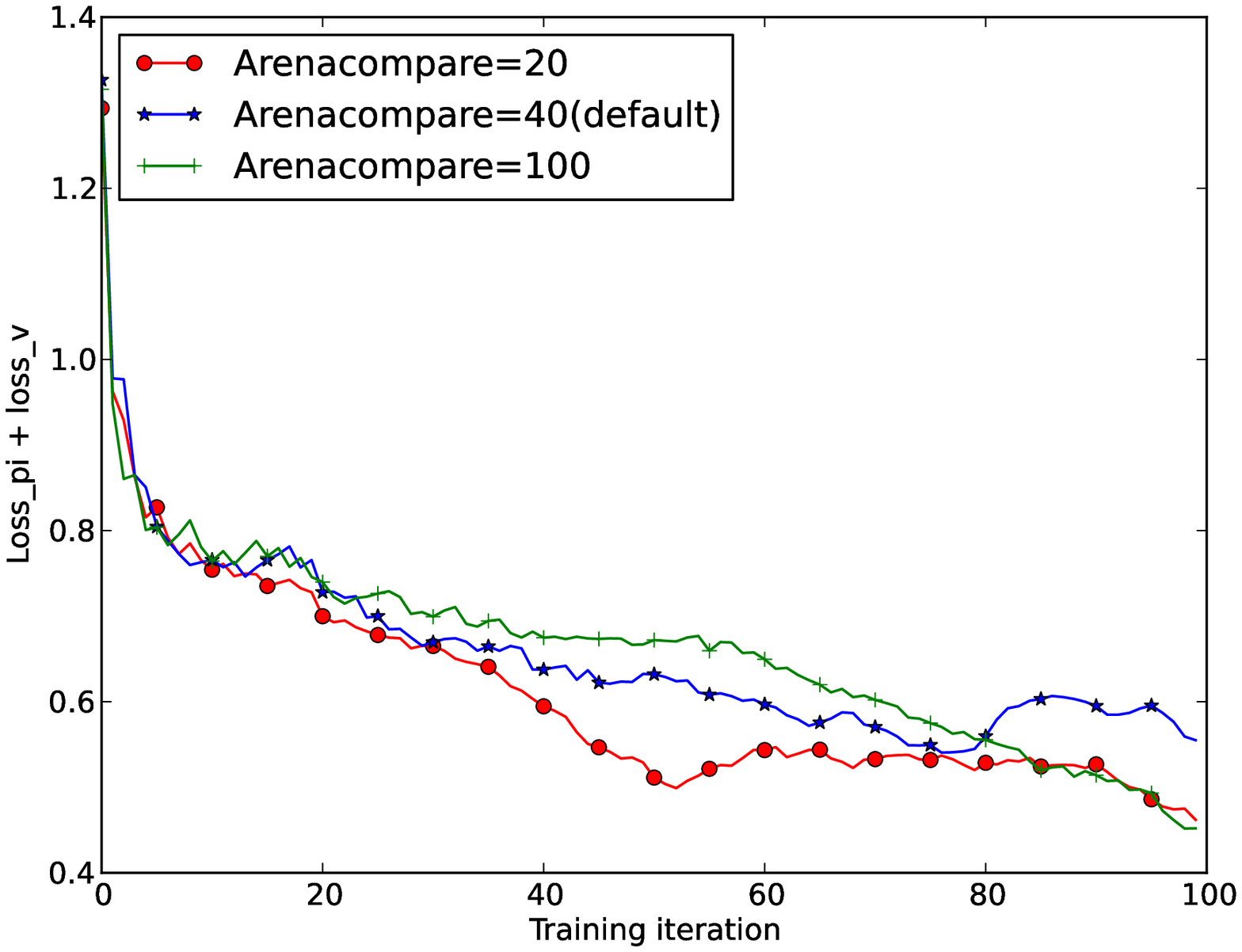}}
\hspace*{-2.3em}
\subfigure[Elo Rating]{\label{fig:subfigarenacompare:b} %% label for second subfigure
\includegraphics[width=0.53\textwidth]{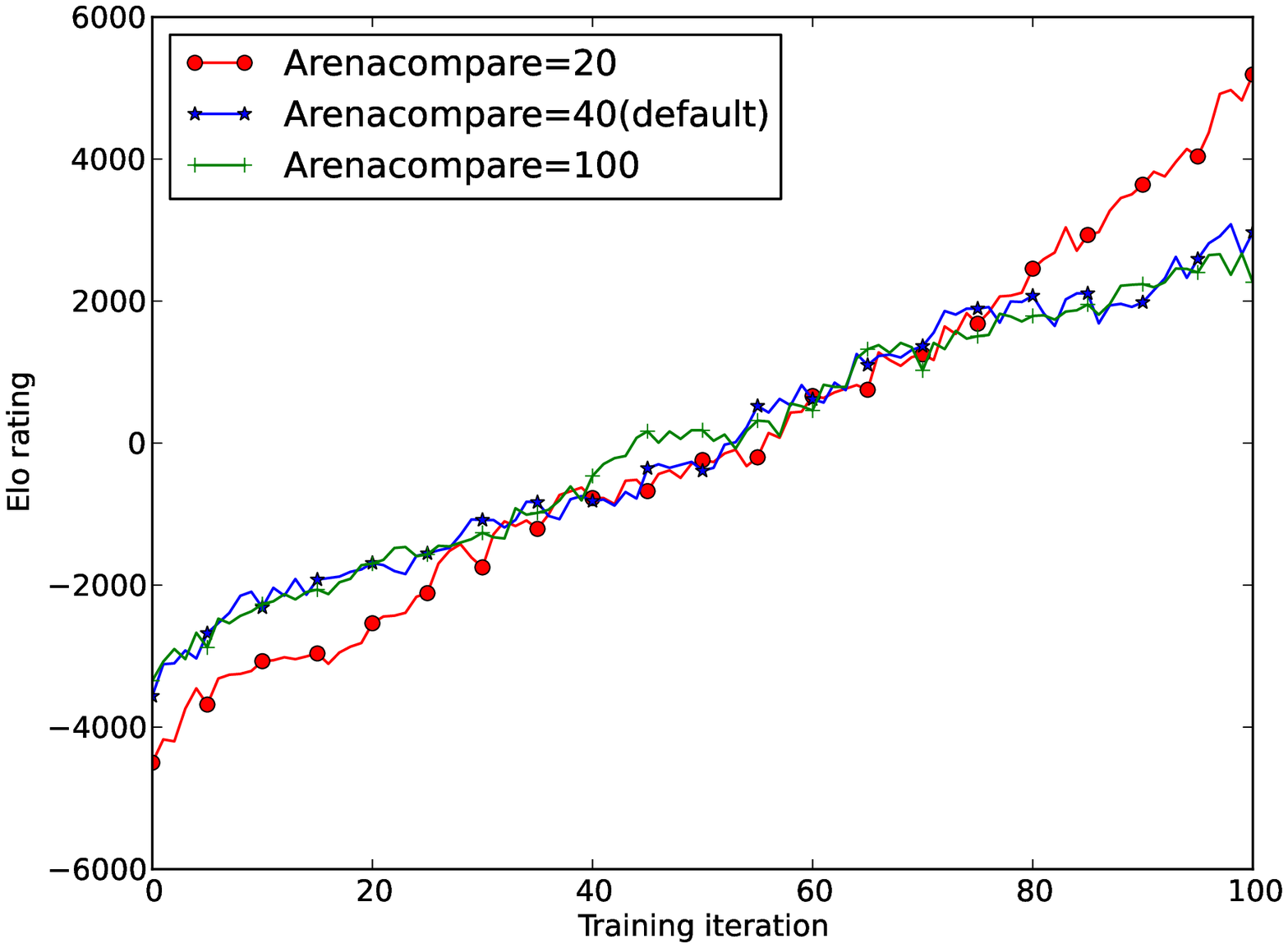}}
%\hspace*{-2.3em}
%\subfigure[Time Cost]{\label{fig:subfigarenacompare:c} %% label for second subfigure
%\includegraphics[width=0.53\textwidth]{time_arenacompare.eps}}
\caption{Training Loss and Elo Rating with Different Arenacompare}
\label{fig:subfigarenacompare} %% label for entire figure
\end{figure}

\textbf{\emph{updateThreshold:}} Normally, in two-player games, the condition of judging A player is better than B player is that A must win more than 50\% games. In order to avoid coincidence, a higher win rate condition is helpful. However, if we set a too high win rate, then it is more difficult to iteratively update to better models. Fig.~\ref{fig:subfigupdateThreshold:a} shows that updateThreshold=0.7 is too high so that the model gets better too slowly than others, which is evidenced by the highest training loss curve. From Fig.~\ref{fig:subfigupdateThreshold:b}, we find that updateThreshold=0.6 is the best value to achieve the highest elo rating.
\begin{figure}[H]
\centering
\hspace*{-2.3em}
\subfigure[Training Loss]{\label{fig:subfigupdateThreshold:a} %% label for first subfigure
\includegraphics[width=0.53\textwidth]{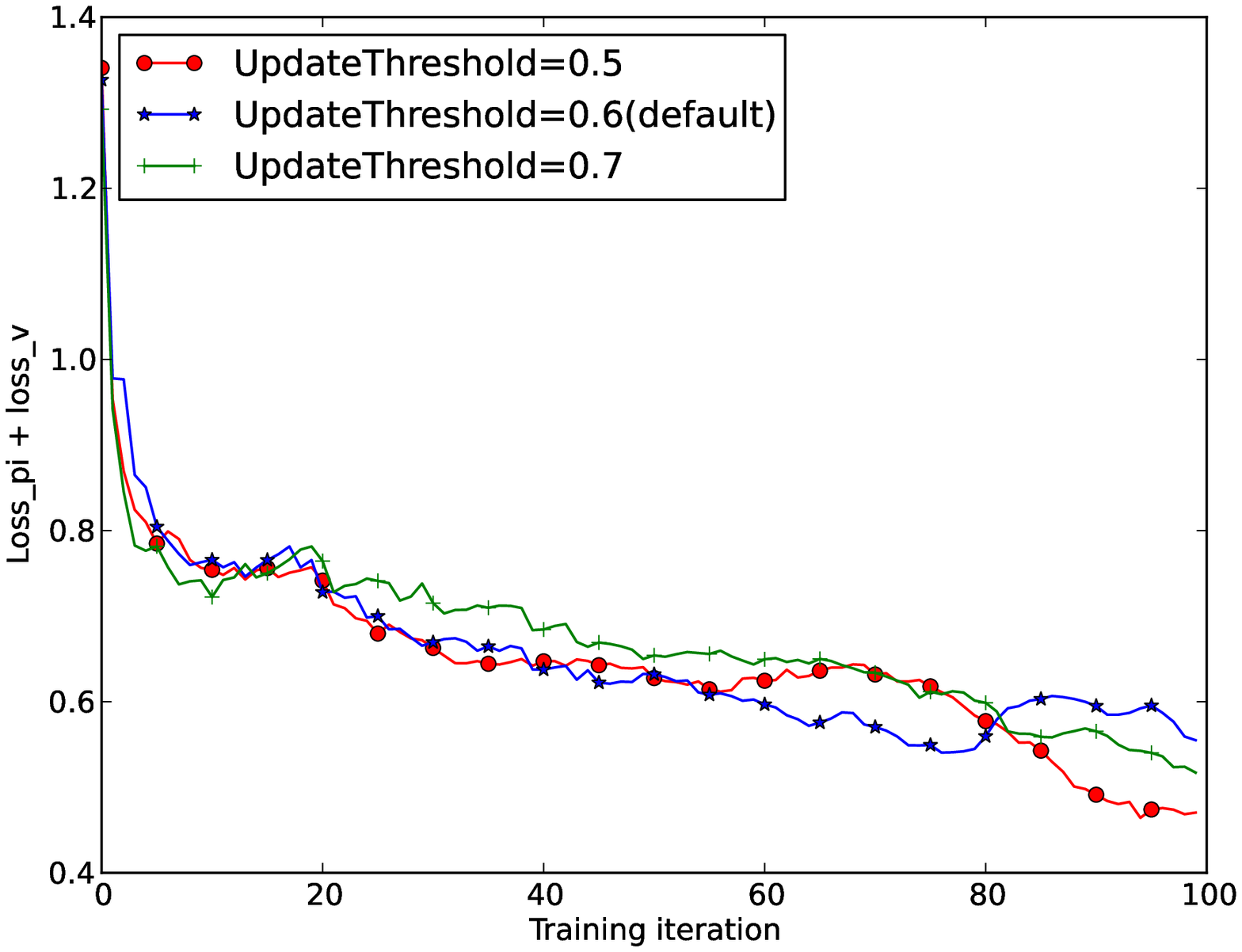}}
\hspace*{-2.3em}
\subfigure[Elo Rating]{\label{fig:subfigupdateThreshold:b} %% label for second subfigure
\includegraphics[width=0.53\textwidth]{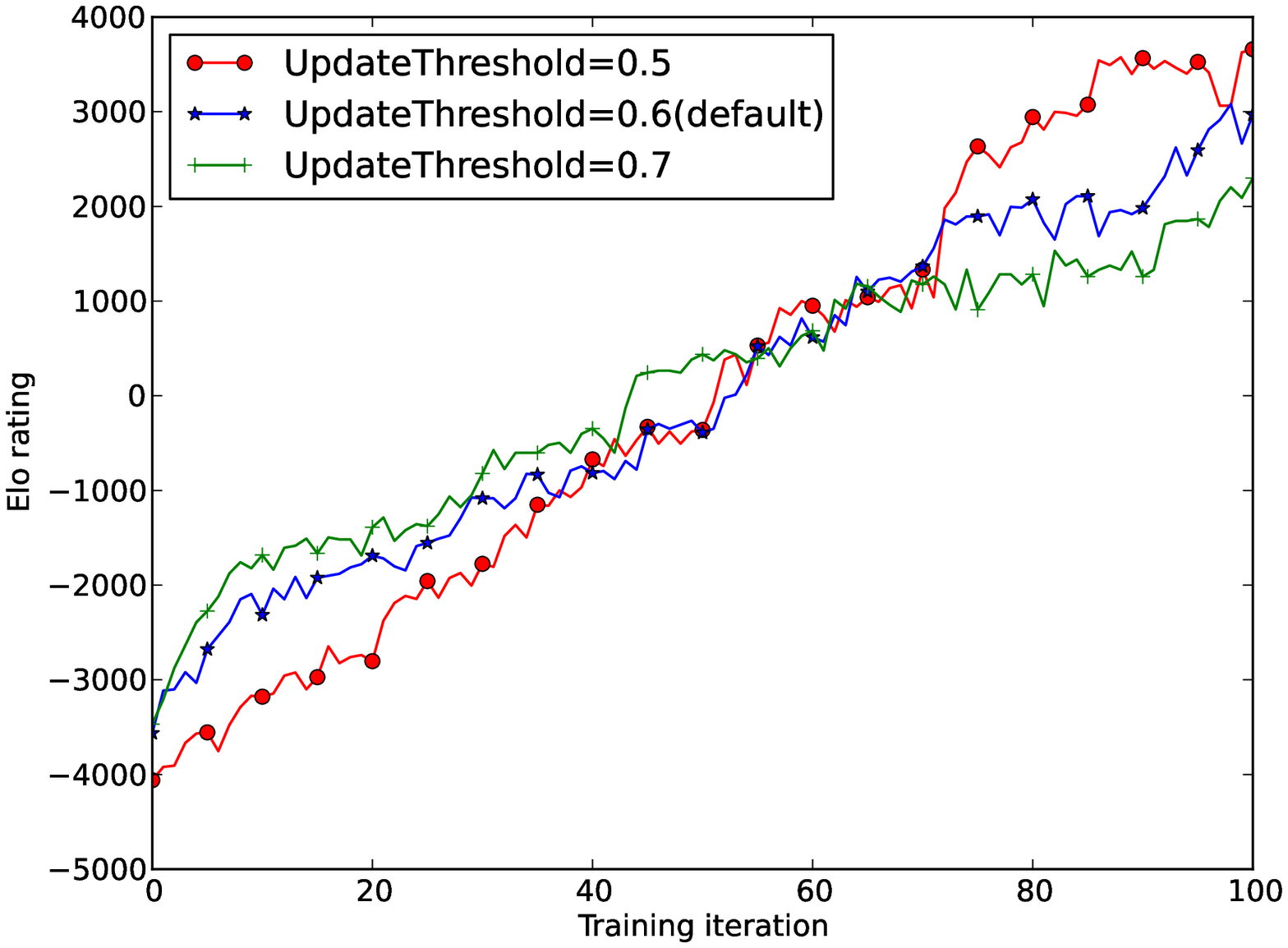}}
%\hspace*{-2.3em}
%\subfigure[Time Cost]{\label{fig:subfigupdateThreshold:c} %% label for second subfigure
%\includegraphics[width=0.53\textwidth]{time_updateThreshold.eps}}
\caption{Training Loss and Elo Rating with Different UpdateThreshold}
\label{fig:subfigupdateThreshold} %% label for entire figure
\end{figure}
We present the final time cost of each experiment using different parameter values in Table~\ref{timecosttab}. From the table, for parameter iteration, episode, mctssimu, retrainlength, arenacompare, smaller values lead to less time cost, which is in expectation. For batchsize, bigger value results to less time cost. The rest parameters are time-friendly, changing their values will not lead to significant different time cost. Therefore, tuning these time-friendly parameters could be effective.
\begin{table}[H]
\centering\hspace*{-2.3em}
%\linebreak
\caption{Time Cost~(hr) of Different Parameter Setting}\label{timecosttab}
\begin{tabular}{|l|l|l|l|l|}
\hline
Parameter	& Minimum Value	& Default Value& Maximum Value& Type\\
\hline
iteration	&\textbf{23.8}	&44.0	&60.3&time-sensitive\\
\hline
episode	&\textbf{17.4}	&44.0&87.7&time-sensitive\\
\hline
tempThreshold	&41.6		&44.0&40.4&time-friendly\\
\hline
mctssimu	&\textbf{26.0}	&44.0&	64.8&time-sensitive\\
\hline
Cpuct	&50.7	&44.0&	49.1&time-friendly\\
\hline
retrainlength	&\textbf{26.5}	&44.0&50.7&time-sensitive\\
\hline
epoch	&\textbf{43.4}	&44.0	&55.7&time-sensitive\\
\hline
batchsize	&47.7	&44.0&\textbf{37.7}&time-sensitive\\
\hline
learningrate	&47.8&	44.0&40.3&time-friendly\\
\hline
dropout&	51.9&44.0&51.4&time-friendly\\
%\hline
%channel&	44.1&	44.0&70.9\\
\hline
arenacompare	&\textbf{33.5}&44.0	&57.4&time-sensitive\\
\hline
updateThreshold	&39.7	&44.0 &	40.4&time-friendly\\
\hline
\end{tabular}
\end{table}
Based on the aforementioned experimental results and corresponding analysis, we summarize the importance by evaluating contributions of each parameter to training loss, elo rating and time cost, respectively, in Table~\ref{losstimetab}. For training loss, different value of \emph{arenacompare} and \emph{updateThreshold} can not make significant difference. For elo rating, different value of \emph{arenacompare} can not make significant difference. For time cost, different value of time-friendly parameters contribute almost the same to the whole training, while usually, bigger value of time-sensitive parameters requires more time to the whole training.
\begin{table}[H]
\centering\hspace*{-2.3em}
%\linebreak
\caption{A Summary of Importance in Different Dimensions}\label{losstimetab}
\begin{tabular}{|l|l|l|l|l|}
\hline
Parameter	&Default Value &	Loss  &Elo Rating& Time Cost\\
\hline
iteration	& 100&100&100 &\textbf{50}\\
\hline
episode	& 50 &\textbf{10}&\textbf{10}&\textbf{10}\\
\hline
tempThreshold	&15 &\textbf{10}&\textbf{10}&similar\\
\hline
mctssimu	& 100&\textbf{200}&\textbf{200}&\textbf{25}\\
\hline
Cpuct	& 1.0&\textbf{0.5}&\textbf{0.5(2.0)}&similar\\
\hline
retrainlength	&20 &\textbf{1}&\textbf{40}&\textbf{1}\\
\hline
epoch	& 10&\textbf{15}&\textbf{15}&\textbf{5}\\
\hline
batchsize	& 64&\textbf{96}&\textbf{32}&\textbf{96}\\
\hline
learningrate &0.005 &\textbf{0.001}&\textbf{0.001}&similar\\
\hline
dropout   &0.3 &0.3&\textbf{0.2}&similar\\
\hline
arenacompare	& 40&insignificant&-&\textbf{20}\\
\hline
updateThreshold	&0.6&insignificant&0.6&similar\\
\hline
\end{tabular}
\end{table}

\section{Conclusion}
In this paper, since there is few work discussed how to set the parameters values in AlphaGo series algorithms. Based on AlphaZero, we analyze 12 parameters in this general framework. First, we introduce the roles and functions of every parameters by showing these parameters' particular places appeared within the algorithm framework. Second, we introduce three objectives~(i.e. training loss function, time cost function and elo rating function) to optimize. Based on an implementation of AlphaZero from github~\cite{Surag2018}, we configure the algorithm with our default parameter setting to do comparison experiments based on 6$\times$6 Othello. 1) We classified time-friendly parameters and time-sensitive parameters, see Table~\ref{timecosttab}. 2) In addition, we find that although some time-sensitive parameters, such as \emph{mctssimulation}, \emph{iteration} and \emph{episode} etc, can make significant difference with different values. Interestingly, not always bigger values result to positive significant improvement, which proves that there exists the optimal value for a given time cost for these parameters. 3). For time-friendly parameters, different values will not make significant difference of time cost, so tuning these parameters can be regarded as a Pareto Improvement process. Our these findings support the applications of AlphaZero and provide us a basis to optimize parameter setting values and transfer the optimal parameter setting values to different games and tasks within AlphaZero framework.

In the future, we can apply automatically optimization framework~\cite{Birattari2002,Hutter2011} techniques to replace manually optimizing parameter in this paper. In addition, relationship among parameters may be much more complicate. Therefore, it is worthy to digging out the co-efficiency. What's more, transfer learning in AlphaZero still has much work to do for us. For instance, we can also transfer models to different tasks to get preliminary and abstract knowledge within AlphaZero framework.

\subsubsection*{Acknowledgments.} Hui Wang acknowledges financial support from the China Scholarship Council (CSC), CSC No.201706990015.


\begin{thebibliography}{22}

\bibitem{Silver2016} Silver D, Huang A, Maddison C J, et al: Mastering the game of Go with deep neural networks and tree search. Nature \textbf{529}(7587), 484--489 (2016)

\bibitem{Silver2017a} Silver D, Schrittwieser J, Simonyan K, et al: Mastering the game of go without human knowledge. Nature \textbf{550}(7676), 354--359 (2017)

\bibitem{Silver2017b} Silver D, Hubert T, Schrittwieser J, et al: Mastering Chess and Shogi by Self-Play with a General Reinforcement Learning Algorithm. arXiv preprint arXiv:1712.01815, (2017).

\bibitem{Granter2017} Granter S R, Beck A H, Papke Jr D J: AlphaGo, deep learning, and the future of the human microscopist. Archives of pathology $\&$ laboratory medicine \textbf{141}(5), 619--621 (2017)

\bibitem{Wang2016} Wang F Y, Zhang J J, Zheng X, et al: Where does AlphaGo go: From church-turing thesis to AlphaGo thesis and beyon. IEEE/CAA Journal of Automatica Sinica \textbf{3}(2), 113--120 (2016)

\bibitem{Fu2016} Fu M C: AlphaGo and Monte Carlo tree search: the simulation optimization perspective. Proceedings of the 2016 Winter Simulation Conference. IEEE Press pp. 659--670 (2016)

\bibitem{Tao2016} Tao J, Wu L, Hu X: Principle Analysis on AlphaGo and Perspective in Military Application of Artificial Intelligence. Journal of Command and Control \textbf{2}(2), 114--120 (2016)

\bibitem{Zhang2016} Zhang Z: When doctors meet with AlphaGo: potential application of machine learning to clinical medicine. Annals of translational medicine \textbf{4}(6), (2016)

\bibitem{Iwata1994} Iwata S, Kasai T. The Othello game on an n$\times$n board is PSPACE-complete. Theoretical Computer Science. \textbf{123}(2), 329--340 (1994)

\bibitem{Wang2018} Wang H, Emmerich M, Plaat A. Monte Carlo Q-learning for General Game Playing. arXiv preprint arXiv:1802.05944 (2018)

\bibitem{Browne2012} Browne C B, Powley E, Whitehouse D, et al: A survey of monte carlo tree search methods. IEEE Transactions on Computational Intelligence and AI in games \textbf{4}(1), 1--43 (2012)

\bibitem{Ruijl2014} B Ruijl, J Vermaseren, A Plaat, J Herik: Combining Simulated Annealing and Monte Carlo Tree Search for Expression Simplification. In: B\'eatrice Duval, H. Jaap van den Herik, St\'ephane Loiseau, Joaquim Filipe. Proceedings of the 6th International Conference on Agents and Artificial Intelligence 2014, vol. 1, pp. 724--731. SciTePress, Set\'ubal, Portugal (2014)

\bibitem{Schmidhuber2015} Schmidhuber J: Deep learning in neural networks: An overview. Neural networks \textbf{61} 85--117 (2015)

\bibitem{Clark2015} Clark C, Storkey A. Training deep convolutional neural networks to play go. International Conference on Machine Learning. pp. 1766--1774 (2015)

\bibitem{Mnih2015} Mnih V, Kavukcuoglu K, Silver D, et al: Human-level control through deep reinforcement learning. Nature \textbf{518}(7540), 529--533 (2015)

\bibitem{Heinz2000} Heinz E A: New self-play results in computer chess. International Conference on Computers and Games. Springer, Berlin, Heidelberg. pp. 262--276 (2000)

\bibitem{Wiering2010} Wiering M A: Self-Play and Using an Expert to Learn to Play Backgammon with Temporal Difference Learning. Journal of Intelligent Learning Systems and Applications \textbf{2}(2), 57--68 (2010)

\bibitem{Van2013} Van Der Ree M, Wiering M: Reinforcement learning in the game of Othello: Learning against a fixed opponent and learning from self-play. In Adaptive Dynamic Programming And Reinforcement Learning. pp. 108--115 (2013)

%\bibitem{Pan2008} Pan S J, Kwok J T, Yang Q: Transfer Learning via Dimensionality Reduction. AAAI Volume 8. pp. 677--682 (2008).

%\bibitem{Pan2010} Pan S J, Yang Q: A survey on transfer learning. IEEE Transactions on knowledge and data engineering \textbf{22}(10), 1345--1359 (2010)

%\bibitem{Taylor2009} Taylor M E, Stone P: Transfer learning for reinforcement learning domains: A survey. Journal of Machine Learning Research \textbf{10}(Jul), 1633--1685 (2009)

%\bibitem{Mishkin2015} Mishkin D, Matas J: All you need is a good init. arXiv preprint arXiv:1511.06422 (2015)

%\bibitem{French1999} French R M. Catastrophic forgetting in connectionist networks. Trends in cognitive sciences \textbf{3}(4), 128--135 (1999)

%\bibitem{Rusu2016} Rusu A A, Rabinowitz N C, Desjardins G, et al: Progressive neural networks. arXiv preprint arXiv:1606.04671 (2016)

\bibitem{Kingma2014} Kingma D P, Ba J: Adam: A method for stochastic optimization. arXiv preprint arXiv:1412.6980 (2014)

\bibitem{Ioffe2015} Ioffe S, Szegedy C: Batch normalization: accelerating deep network training by reducing internal covariate shift. Proceedings of the 32nd International Conference on International Conference on Machine Learning-Volume 37. pp. 448--456 (2015)

\bibitem{Srivastava2014} Srivastava N, Hinton G, Krizhevsky A, et al: Dropout: a simple way to prevent neural networks from overfitting. The Journal of Machine Learning Research. \textbf{15}(1), 1929--1958 (2014)

\bibitem{Albers2001} Albers P C H, Vries H. Elo-rating as a tool in the sequential estimation of dominance strengths. Animal Behaviour 489--495 (2001)

\bibitem{Coulom2008} Coulom R. Whole-history rating: A Bayesian rating system for players of time-varying strength. International Conference on Computers and Games. Springer, Berlin, Heidelberg, 113--124, 2008

\bibitem{Schneider2002} Schneider M O, Rosa J L G: Neural connect 4-A connectionist approach to the game. Neural Networks SBRN 2002. Proceedings. VII Brazilian Symposium on. IEEE pp. 236--241 (2002)

\bibitem{Surag2018} Surag Nair, https://github.com/suragnair/alpha-zero-general

\bibitem{Birattari2002} Birattari M, St\"utzle T, Paquete L, et al. A racing algorithm for configuring metaheuristics. Proceedings of the 4th Annual Conference on Genetic and Evolutionary Computation. Morgan Kaufmann Publishers Inc. 11-18 (2002)

\bibitem{Hutter2011} Hutter F, Hoos H H, Leyton-Brown K: Sequential model-based optimization for general algorithm configuration. International Conference on Learning and Intelligent Optimization. Springer, Berlin, Heidelberg, pp. 507--523 (2011)
\end{thebibliography}
\end{document}